\documentclass[journal]{IEEEtran}

\usepackage{amsmath,amsfonts,amssymb,amsthm,bbm,cite,graphicx,makecell,mathrsfs,mathtools,pgfplots,stmaryrd,tikz,units}
\pgfplotsset{compat=1.3}
\usepackage{caption}
\usepackage{subcaption}
\usepackage{multicol}
\usepackage{centernot}

\newtheorem{definition}{Definition}

\newtheorem{assumption}{Assumption}
\usepackage[ruled,vlined]{algorithm2e}
\usepackage{float}
\usepackage{dblfloatfix}

\usepackage{pgfplots}
\usepgfplotslibrary{dateplot}

\usepackage[hidelinks]{hyperref}

\newcommand{\overbar}[1]{\mkern 1.5mu\overline{\mkern-1.5mu#1\mkern-1.5mu}\mkern 1.5mu}

\newcommand{\norm}[1]{\left\lVert#1\right\rVert}
\newcommand{\evaltzero}{{\Bigr\rvert_{t = 0}}}

\newcommand{\Det}[1]{\left|{#1}\right|} % Determinant Operator
\DeclareMathOperator*{\argmin}{arg\,min}

\DeclareMathOperator{\card}{card}

\DeclareMathOperator{\D}{D}
\DeclareMathOperator{\diag}{\textup{diag}}

\DeclareMathOperator{\grad}{grad}
\DeclareMathOperator{\sym}{sym}

\DeclareMathOperator*{\Tr}{Tr}

\DeclareMathOperator*{\minimize}{minimize}

\DeclareMathOperator{\KL}{KL}

\newcommand{\N}{\mathcal{N}}

\newcommand{\E}{\mathbb{E}}

\newcommand*{\VEC}[1]  {{\ensuremath{\boldsymbol{#1}}}}
\newcommand*{\MAT}[1]  {{\ensuremath{\boldsymbol{#1}}}}
\newcommand{\ba}{\VEC{a}}
\newcommand{\bA}{\MAT{A}}

\newcommand{\bB}{\MAT{B}}

\newcommand{\bEta}{\VEC{\eta}}

\newcommand{\G}{\MAT{G}}
\newcommand{\bI}{\MAT{I}}

\newcommand{\bL}{{\MAT{\Lambda}}}
\newcommand{\bMu}{{\VEC{\mu}}}
\newcommand{\bMuG}{{\bMu^{\textup{G}}}}
\newcommand{\bMuIG}{{\bMu^{\textup{IG}}}}
\newcommand{\bMuT}{{\bMu^{\textup{Ty}}}}

\newcommand{\bNu}{{\MAT{\nu}}}

\newcommand{\bQ}{\MAT{Q}}

\newcommand{\bSigma}{{\MAT{\Sigma}}}
\newcommand{\bSigmaG}{{\bSigma^{\textup{G}}}}
\newcommand{\bSigmaGcentered}{{\bSigma^{\textup{G}, \bMu=\VEC{0}}}}
\newcommand{\bSigmaIG}{{\bSigma^{\textup{IG}}}}
\newcommand{\bSigmaT}{{\bSigma^{\textup{Ty}}}}
\newcommand{\bSigmaTmu}{{\bSigma^{\textup{Ty}, \bMu}}}

\newcommand{\bt}{{\boldsymbol{\tau}}}

\newcommand{\bu}{{\VEC{u}_i}}
\newcommand{\U}{{\MAT{U}}}

\newcommand{\bx}{{\VEC{x}}}
\newcommand{\bX}{\MAT{X}}
\newcommand{\bXi}{\VEC{\xi}}
\newcommand{\bY}{\MAT{Y}}
\newcommand{\0}{\MAT{0}}
\newcommand{\1}{\mathbbm{1}_n}

\newcommand{\rdot}{{\dot{r}}}
\newcommand{\rddot}{{\ddot{r}}}
\newcommand{\bxt}{{\VEC{x}(t)}}
\newcommand{\bxtdot}{{\dot{\VEC{x}}(t)}}
\newcommand{\bxtddot}{{\ddot{\VEC{x}}(t)}}

\newcommand{\bMudot}{{\dot{\bMu}}}
\newcommand{\btdot}{{\dot{\bt}}}
\newcommand{\bSigmadot}{{\dot{\bSigma}}}
\newcommand{\bMuddot}{{\ddot{\bMu}}}
\newcommand{\btddot}{{\ddot{\bt}}}
\newcommand{\bSigmaddot}{{\ddot{\bSigma}}}

\newcommand{\R}{\mathbb{R}}

\newcommand{\Eset}{{\mathcal{E}}}
\newcommand{\Emsg}{{\Eset_{p,n}}}

\newcommand{\Sym}{{\mathcal{S}_p}}

\newcommand{\Spos}{{\mathcal{S}_p^{++}}}

\newcommand{\Mpn}{{\mathcal{M}_{p,n}}}
\newcommand{\M}{{\mathcal{M}}}
\newcommand{\Mg}{{\mathcal{M}_p}}
\newcommand{\Mmsg}{{\mathcal{M}_{p,n}}}

\newcommand{\Op}{\mathcal{O}_p}

\newcommand{\RPos}{{\R_*^+}}
\newcommand{\RPosExt}{{\overbar{\R^+}}}
\newcommand{\RPosVec}{{(\R_*^+)^n}}

\newcommand{\Skew}{\mathcal{A}_p}
\newcommand{\SRPosVec}{{\mathcal{S}(\R_*^+)^n}}

\newtheorem{proposition}{Proposition}

\newcommand{\NLL}{{\mathcal{L}}}
\newcommand{\NLLReg}{\NLL_\Reg}
\newcommand{\Reg}{{\mathcal{R}_\kappa}}
\newcommand{\reg}{{r_\kappa}}

\begin{document}
\title{
	Riemannian optimization for non-centered mixture of scaled Gaussian distributions
	% Non-centered mixture of scaled Gaussian distributions
	% Non-centered mixture of scaled Gaussian distributions: blabla
}
\author{Antoine Collas, Arnaud Breloy, Chengfang Ren, Guillaume Ginolhac, Jean-Philippe Ovarlez
	\thanks{
	A. Collas, C. Ren are with SONDRA, CentraleSupélec, Université Paris-Saclay.
	G. Ginolhac is with LISTIC (EA3703), University Savoie Mont Blanc.
	A. Breloy is with LEME (EA4416), University Paris Nanterre.
	J.-P. Ovarlez is with SONDRA, CentraleSupélec, Université Paris-Saclay and DEMR, ONERA, Université Paris-Saclay.
	% Part of this work was supported by ANR-ASTRID MARGARITA (ANR-17-ASTR-0015). % faut laisser margarita encore ?
	Part of this work was supported by ANR MASSILIA (ANR-21-CE23-0038-01).
	}
}

\maketitle

\begin{abstract}
	This paper studies the statistical model of the non-centered mixture of scaled Gaussian distributions (NC-MSG).
	Using the Fisher-Rao information geometry associated with this distribution, we derive a Riemannian gradient descent algorithm.
	This algorithm is leveraged for two minimization problems.
	The first is the minimization of a regularized negative log-likelihood (NLL).
	The latter makes the trade-off between a white Gaussian distribution and the NC-MSG.
	Conditions on the regularization are given so that the existence of a minimum to this problem is guaranteed without assumptions on the samples.
	Then, the Kullback-Leibler (KL) divergence between two NC-MSG is derived.	
	This divergence enables us to define a second minimization problem.
	The latter is the computation of centers of mass of several NC-MSGs.
	% The proposed Riemannian gradient descent algorithm is leveraged to solve this second minimization problem.
	Numerical experiments show the good performance and the speed of the Riemannian gradient descent on the two problems.
	Finally, a \emph{Nearest centroïd classifier} is implemented leveraging the KL divergence and its associated center of mass.
	Applied on the large-scale dataset \emph{Breizhcrops}, this classifier shows good accuracies and robustness to rigid transformations of the test set.
\end{abstract}


\section{Introduction}
	\label{sec:intro}
	The first and second-order statistical moments of the sample set $\{\bx_i\}_{i=1}^n \in (\R^p)^n$ are ubiquitous features in signal processing and machine learning algorithms.
Classically, these parameters are estimated using the empirical mean and the sample covariance matrix (SCM), which correspond to the maximum likelihood estimators (MLE) of the multivariate Gaussian model.
However, these estimates tend to perform poorly in the context of heavy-tailed distributions, or when the sample set contains outliers.
In such setups, one can obtain a better fit to empirical distributions by considering more general statistical models, such as the elliptical distributions~\cite{KY90}.
Within this broad family of distributions, $M$-estimators of the location and scatter~\cite{M76} appear as generalized MLEs and have been leveraged for their robustness properties in many fields (cf.~\cite{OTKP12} for an extensive review).

An important subfamily of elliptical distributions is the compound Gaussian distributions, which models samples as $\bx \overset{d}{=} \bMu + \sqrt{\tau} \bu $, 
where $\bMu \in \R^p$ is the center (also referred to as location) of the distribution, $\bu \sim \N(\0, \bSigma)$ is the \textit{speckle} (centered Gaussian distribution with covariance matrix $\bSigma$), and $\tau\in\RPos$ is an independent random scaling factor called the \textit{texture}.
The flexibility regarding the choice of the probability density function for $\tau$ results in various models for $\bx$.
Compound Gaussian distributions encompass the $t$-distribution (including the Cauchy distribution), and the $K$-distribution.
In practice, the underlying distribution is generally unknown, which is why the textures have often been modeled as unknown and deterministic in the centered case, \emph{i.e.}, $\bx_i \sim \N(\0, \tau_i \bSigma)$.
Such model will be referred to as \textit{mixture of scaled Gaussian distributions} (MSG)~\cite{W11}.
The MLE of the scatter matrix $\bSigma$ of this model coincides with Tyler's $M$-estimator of the scatter~\cite{T87}, which attracted considerable activity due to its robustness and distribution-free properties over the elliptical distributions family~\cite{CMR02, PFOL08, FJ08, ZCS16}.
However, its transposition to the non-centered case from the model $\bx_i \sim \N(\bMu, \tau_i \bSigma)$ received less interest.
%\footnote{ Notice that D. E. Tyler also proposes an $M$-estimator of location and scatter that is a solution of a fixed point equation in~\cite{T87}.
%While the MLE of $\bx_i \sim \N(\0, \tau_i \bSigma)$ and Tyler's estimator coincide for the scatter matrix, this is not the case for the non-centered model~\cite{CBBRGO21}.
%}
This might be because the usual fixed-point iterations to evaluate its maximum likelihood may diverge in practice, which motivated the present work.

In this paper, we tackle optimization problems related to parameter estimation and classification for a non-centered mixture of scaled Gaussian distributions (NC-MSG).
The contribution is threefold:

First, we derive a new Riemannian gradient descent algorithm based on the Fisher-Rao information geometry of the considered statistical model.
Indeed the parameter space (location, scatter, textures) is a product manifold that can be endowed with a Riemannian metric to leverage the Riemannian optimization framework~\cite{AMS08, B22}.
The Fisher-Rao information geometry corresponds to the one induced by the Fisher information metric.
It is of particular interest since it is inherently well suited to the natural geometry of the data~\cite{N22}.
In this scope, we derive the Riemannian gradient (also referred to as the natural gradient) and a second-order retraction of this geometry.
These tools are enough to cast a gradient descent applicable to any function of the parameters.
We focus on two prominent examples that are regularized maximum likelihood estimation and center of mass computation.
Simulations evidence that the proposed approach allows for fast computation of critical points, as it can converge with up to one order of magnitude less of iterations compared to other Riemannian descent approaches.

The second line of contributions concerns the problem of maximum likelihood estimation, for which we propose a new class of regularization penalties.
A main issue with NC-MSGs is that the existence of the maximum likelihood is not guaranteed.
This is due to attraction points where the likelihood function diverges.
This also explains why standard fixed-point algorithms to evaluate the solution may diverge in practice.
Related issues are well known in the context of $M$-estimators because their existence is subject to strict conditions that are not always met in practice~\cite{M76, T87, OTKP12}, for example when there is insufficient sample support ($n<p$).
In such setups,  it is common to rely on regularization penalties to ensure the existence of a solution, and the stability of corresponding iterative algorithms.
In the centered case of elliptical distributions, several works considered shrinkage of $M$-estimators to a target scatter matrix~\cite{PCQ14, SBP14,OT14}, and regularizing both the mean and the scatter for the non-centered $t$-distribution was studied in~\cite{SBP15}.
Other regularizations formulated on the spectrum of the scatter matrix were proposed in~\cite{W11, BOP19, YT20} for the centered case.
For NC-MSGs, we propose here a family of penalties that can be interpreted as a divergence between the initial model and a white Gaussian one (\emph{i.e.}, that shrinks both the textures and eigenvalues of the scatter matrix to a pre-defined $\kappa \in \RPos$).
We derive the general conditions for these penalties to ensure the existence of a solution for the regularized MLE.
Interestingly, we show that this existence is only conditioned to the design of the penalty, and does not depend on the size of the sample support.
We also study the invariance properties of the resulting estimators.

Finally, we apply the proposed algorithm to perform Riemannian classification.
We consider the framework where statistical features of sample batches are used to discriminate between classes~\cite{BBCJ12, TPM07, TPM08, FOP13}.
The Riemannian approach then consists in generalizing usual classification algorithms (\emph{e.g.}, the \emph{Nearest centroïd classifier}) by replacing the Euclidean distance and arithmetic mean by divergence and its corresponding center of mass~\cite{K77, ABY13, CKBM21}.
In this setup, the information geometry can help design meaningful distances between the features, and improve the output performance~\cite{BBCJ12, FOP13}.
Unfortunately, the geodesic distance associated with the Fisher information metric of the NC-MSG remains unobtainable in closed-form (it is still unknown for the non-centered multivariate Gaussian model~\cite{CO91, TRZL19, CBGBRO22}).
Instead, we propose to rely on the Kullback-Leibler (KL) divergence and its associated center of mass (computed using the proposed Riemannian optimization algorithm).
We apply such Riemannian classification framework to the \emph{Breizhcrops} dataset~\cite{breizhcrops2020}.
% Our experiments evidence that using the whole parameters of the NC-MSG model can improve the classification process compared to classifying with individual features only.
Our experiments evidence that regularizing the estimation greatly improves the accuracy.
Thanks to the invariance properties of the proposed estimators, we also show that this process exhibits good robustness to rigid transformations of the samples during the inference.

The rest of the paper is organized as follows.
Section~\ref{sec:model} presents NC-MSGs and casts their parameter space as a manifold.
Section~\ref{sec:Riemannian_geometry} presents elements of Riemannian geometry, and studies the Fisher-Rao information geometry for this model. 
Section~\ref{sec:Riemannian_opt} derives a Riemannian gradient descent algorithm following this geometry.
Section~\ref{sec:MLE} discusses parameter estimation in the considered model, presents a new class of regularized estimators, and studies some of their properties (existence, invariances).
Section~\ref{sec:ML} derives the KL divergence of the model and its associated center of mass.
Section~\ref{sec:num_exp} concludes with validation simulations and an application to Riemannian classification of the \emph{Breizhcrops} dataset.
For conciseness, some technical proofs are in appendices that are provided as supplementary materials.

\section{Non-centered mixture of scaled Gaussian distributions and its parameter space \texorpdfstring{$\Mmsg$}{Mmsg}}
	\label{sec:model}
	\subsection{Data model}

Let a set of $n$ data points $\{\bx_i\}_{i=1}^n$ belonging to $\R^p$ and distributed according to the following statistical model
\begin{equation}
	\label{eq:model}
	\bx_i \overset{d}{=} \bMu + \sqrt{\tau_i} \,\bSigma^{\frac12} \, \bu \, ,
\end{equation}
where $\bu$ follows a white circular Gaussian distribution \emph{i.e.} $\bu \sim \N(\0, \bI_p)$.
The variables $\bMu \in \R^p$ and $\bSigma \in \Spos$ (set of $p \times p$ symmetric positive definite matrices) are respectively named the location and scatter parameters.
Then, the unknown texture parameters $\{\tau_i\}_{i=1}^n$ are stacked into the vector $\bt \in \RPosVec$ (set of strictly positive vectors).
If these textures admit a probability density function (p.d.f.), then the random variables (r.v.) $\bx_i$ follow a Compound Gaussian distribution~\cite{OTKP12b, OTKP12}.
However, in general, this p.d.f. is unknown.
In order not to rely on additional pdf assumptions on the textures, these are often assumed to be unkown and deterministic~\cite{PCOFL07, PFOL08}.
In this case, the r.v. $\bx_i$ follow a NC-MSG, \emph{i.e.}
\begin{equation}
	\bx_i \sim \N(\bMu, \tau_i \bSigma)\, .
	\label{eq:distribution}
\end{equation}
Thus $\bx_i$ admits a p.d.f. $f$ defined from the Gaussian one $f_G$
\begin{equation}
	f\left(\bx_i | \left(\bMu, \bSigma, \tau_i\right)\right) = f_G\left(\bx_i | \left(\bMu, \tau_i \bSigma\right)\right)\, ,
\end{equation}
with $\forall \bx \in \R^p$
\begin{multline}
	f_G\left(\bx | (\bMu, \bSigma)\right) =\\ (2 \pi)^{-\frac{p}{2}} |\bSigma|^{-\frac12} \exp\left[ -\frac12 (\bx - \bMu)^T \bSigma^{-1} (\bx - \bMu)  \right] .
\end{multline}

The negative log-likelihood (NLL) of the sample set $\{\bx_i\}_{i=1}^n$ is then defined on the set of parameters $\theta = (\bMu, \bSigma, \bt) \in \R^p \times \Spos \times \RPosVec$ as (neglecting terms not depending on $\theta$)
\begin{multline}
	\label{eq:NLL}
	\NLL\left(\theta | \{\bx_i\}_{i=1}^n \right) =\\ \frac12 \! \sum_{i=1}^n \bigg[\! \log \left|\tau_i \bSigma\right| + \frac{(\bx_i - \bMu)^T \bSigma^{-1} (\bx_i - \bMu)}{\tau_i} \bigg] \, .
\end{multline}
One can observe an ambiguity between the textures $\bt$ and the scatter matrix $\bSigma$.
Indeed, $\forall \alpha > 0$, we have
\begin{multline}
	\label{eq:NLL_ind}
	\NLL\left( \left( \bMu, \alpha\bSigma, \alpha^{-1}\bt \right) | \{ \bx_i \}_{i=1}^n \right) =\\ \NLL \left(  \left( \bMu, \bSigma, \bt \right) | \{ \bx_i \}_{i=1}^n  \right).
\end{multline}
Thus, to identify the textures and scatter matrix parameters, a constraint on $\bt$ or $\bSigma$ can be added.
Here the choice is made to constrain the textures by fixing their product to be equal to one, \emph{i.e.} $\prod_{i=1}^n \tau_i = 1$.
We point out that most of the results of the paper could be obtained by constraining the scatter matrix instead of the textures, with a unit determinant constraint, \emph{i.e.} $\Det{\bSigma} = 1$~\cite{BGRB19, BMZSGB20}.
Then, the parameter space of interest is
\begin{equation}
	\Mmsg = \R^p \times \Spos \times \SRPosVec \, ,
\end{equation}
where $\SRPosVec$ is the set of textures with the unit product,
\begin{equation}
	\SRPosVec = \left\{ \bt \in \RPosVec: \prod_{i=1}^n \tau_i = 1 \right\}.
\end{equation}
% The choice of constraining the textures to have a unit product is motivated by two results additional to the identifiability:
The choice of adding a constraint is motivated by two results additional to the identifiability:
\emph{(i)} it reduces the dimension of the parameter space by removing the indeterminacy~\eqref{eq:NLL_ind},
\emph{(ii)} the associated FIM (see Proposition~\ref{prop:FIM}) admits a simpler expression, which will be instrumental in the rest of the paper as it turns $\Mmsg$ into a Riemannian manifold.
Its simple formula could not have been obtained without adding this constraint (either on $\bt$ or its counterpart on $\bSigma$).
 
% Then, the MLE $\theta$ satisfies the following fixed point equations:
% \begin{equation}
% 	\label{eq:FPE}
% 	\left\{\begin{aligned}
% 		&\bMu = \left(  {\displaystyle\sum_{i=1}^n\frac{1}{\tau_i}}\right)^{-1} \sum_{i=1}^n \frac{\bx_i}{\tau_i} \\
% 		&\bSigma = \frac1n \sum_{i=1}^n\frac{(\bx_i - \bMu)(\bx_i - \bMu)^T}{\tau_i}\\
% 		&\tau_i = \frac1p \, (\bx_i - \bMu)^T\bSigma^{-1}(\bx_i - \bMu).
% 	\end{aligned}\right.
% \end{equation}
\subsection{Related works}

When $\{\bx_i\}_{i=1}^n$ is sampled from an underlying heavy-tailed Compound Gaussian distribution, the empirical mean and SCM do not provide robust and accurate estimates of $\boldsymbol{\mu}$ and $\boldsymbol{\Sigma}$.
In this setup, $M$-estimators \cite{M76}, raised increasing interest in the past decades (see e.g. \cite{OTKP12}). These estimators are expressed through the two joint fixed-point equations 
\begin{equation}
	\label{eq:Mestimator}
	\begin{aligned}
		\bMu    & =  \Big(\sum_{i=1}^n u_1(t_i)\Big)^{-1} \sum_{i=1}^n u_1(t_i) \bx  \triangleq \mathcal{H}_{\bMu}(\bMu,\bSigma) \, ,\\
		\bSigma & = \frac1n \sum_{i=1}^n  u_2(t_i) (\bx - \bMu)(\bx - \bMu)^T  \triangleq \mathcal{H}_{\bSigma}(\bMu,\bSigma)\, ,
	\end{aligned} 
\end{equation}
where $t_i \triangleq  (\bx - \bMu)^T \bSigma^{-1} (\bx - \bMu)$, $u_1$ and $u_2$ are functions that respect Maronna's conditions\footnote{Notice that \cite{M76} rather uses a formulation of \eqref{eq:Mestimator} involving ``$u_1(t_i)$" and ``$u_2(t_i^2)$", with $t_i=\sqrt{(\bx - \bMu)^T \bSigma^{-1} (\bx - \bMu)}$. Without loss of generality, this paper uses the present notation to simplify some discussions.} \cite{M76}. Under certain conditions \cite{M76}, these estimators can be computed with fixed-point iterations
\begin{equation}
	\label{eq:Mestimator_algo}
	\begin{aligned}
		&\bMu_{k+1}  = \mathcal{H}_{\bMu}(\bMu_{k},\bSigma_{k}) \, , \\
		&\bSigma_{k+1} =  \mathcal{H}_{\bSigma}(\bMu_{k+1},\bSigma_{k})\, ,\\
		\end{aligned} 
\end{equation}
that converge towards a unique solution satisfying \eqref{eq:Mestimator}.
Interestingly, some $M$-estimators also appear as MLE when $u_1(t)=u_2(t)$ is linked  to the p.d.f. of an elliptical distribution \cite{OTKP12}.
Expressing these estimators as the solution of an optimization problem drove a more recent line of work leveraging optimization theory allowing, e.g., for generalizations to  structured scatter matrix matrices \cite{WZ15,SBP16,MREBF19} or regularized location and scatter matrix estimation \cite{SBP15}.

In the context of scatter matrix estimation, Tyler's $M$-estimator \cite{T87} is especially interesting thanks to its robustness and ``distribution-free'' properties over the elliptical distributions family. 
Tyler's $M$-estimator is obtained for $\bMu=\mathbf{0}$ and $u_2(t) = p/t$, and also coincide with the MLE of the centered MSG \cite{CMR02, PFOL08}.
However, this estimator cannot trivially be transposed to the case of joint mean-scatter matrix estimation.
Indeed, the MLE solution associated with NC-MSG is obtained with $u_1(t)= u_2(t) = p/t$, which does not satisfy Maronna's conditions \cite{M76}, and for which the fixed-point iterations \eqref{eq:Mestimator_algo} generally diverge. 
Thus, Tyler's $M$-estimator of the scatter matrix is usually applied on demeaned data, where the mean is estimated in a prior step\footnote{
We point out that a closely related estimator proposed in \cite{T87} uses $u_1(t)=\sqrt{p/t}$ and $u_2(t) = p/t$, which yields converging fixed-point iterations in practice despite being a limit case of Maronna's conditions. 
This estimate, however, is not obtained as the solution of an underlying optimization problem, i.e., has no MLE interpretation.}.
It was yet experienced that the MLE of NC-MSG could be evaluated in practice with Riemannian optimization rather than potentially unstable fixed-point iterations in \cite{CBBRGO21}
(still, without any theoretical guarantees).
The following of this paper builds upon this finding in several directions:
optimization in sections \ref{sec:IG_compound_Gaussian} and \ref{sec:Riemannian_opt},
regularized estimation with theoretical guarantees in section \ref{sec:MLE},
and classification in sections \ref{sec:ML} and~\ref{sec:num_exp}.

\section{Riemannian geometry of \texorpdfstring{$\Mmsg$}{Mmsg}}
	\label{sec:Riemannian_geometry}
	The objective of this section is to present the information geometry of the NC-MSG~\eqref{eq:distribution}; \emph{i.e.} the Riemannian geometry of $\Mmsg$ with the FIM as a Riemannian metric.
This Riemannian geometry is leveraged to optimize several cost functions $h: \Mmsg \to \R$.
Notably, two cost functions will be studied: a regularized NLL in  Section~\ref{sec:MLE}, and a cost function to compute centers of mass of sets of points $\{ \theta_i \} \subset \Mmsg$ in Section~\ref{sec:ML}.
Before turning $\Mmsg$ into a Riemannian manifold, a brief introduction to Riemannian geometry is made.
For a complete introduction to the topic, see~\cite{AMS08, B22}.

\subsection{Riemannian geometry}
\label{subsec:intro_Riemannian_geometry}

Let $\Eset$ be a \emph{linear space} of dimension $d$.
Informally, a \emph{smooth embedded manifold} $\M \subset \Eset$ of dimension $l \leq d$ is a nonempty set that locally resembles a $l$-dimensional linear space.
Indeed, $\M$ is a smooth embedded manifold of $\Eset$ if and only if it is locally diffeomorphic\footnote{A \emph{diffeomorphism } is a bijective map $f: U \to V$ where $U, V$ are open sets and such that both $f$ and $f^{-1}$ are smooth (or infinitely differentiable).} with open sets of a $l$-dimensional linear subspace in $\R^d$.
Then, \emph{smooth curves} $c$ are smooth functions from open intervals $I$ of $\R$ to $\M$; \emph{i.e.} $c : I \to \M$.
Collecting velocities of the curves passing through $x \in \M$, we get the \emph{tangent space} at $x$:
\begin{equation}
	T_x\M = \left\{ c'(0) \, | \, c: I \to \M \textup{ is smooth and } c(0)=x \right\}.
\end{equation}
This tangent space corresponds to a linearization of $\M$ at $x$.
The tangent bundle of $\M$ is then the disjoint union of all the tangent spaces of $\M$, i.e., 
$T\M = \left\{  
(x,\xi):x\in\M~\text{and}~\xi\in T_x\M
\right\}$.

So far, we have defined the notion of the smooth embedded manifold of a linear space.
To turn $\M$ into a \emph{Riemannian manifold}, its tangent spaces $T_x\M$ are equipped with a \emph{Riemannian metric} which is an \emph{inner product}\footnote{An \emph{inner product} is a bilinear, symmetric, positive definite function on a $\R$-vector space.} $\langle ., .\rangle_x^\M: T_x\M \times T_x\M \to \R$ that varies smoothly with respect to $x$.\footnote{For all smooth vector fields $\xi, \eta$ on $\M$ the function $x \mapsto \langle \xi, \eta \rangle_x^\M$ is smooth.}

Then, to move on $\M$, a \emph{geodesic} is a smooth curve on $\M$ with zero acceleration along its path.
In a \emph{Euclidean space} $\Eset$ the acceleration is classically defined as the second derivative.
Thus, a geodesic $c: I \to \Eset$ is such that $\ddot{\gamma}(t) = 0$ $\forall t \in I$.
If $\gamma(0) = x$ and $\dot{\gamma}(0) = \xi$, then, by integrating, we recover the classical straight line $\gamma(t) = x + t \,\xi$.
This notion of acceleration is generalized to manifolds using the \emph{Levi-Civita connection} denoted by $\nabla$.
This notion requires first defining smooth vector fields, which are smooth mappings that associate a vector in $T\M$ for each point of the manifold $\M$, i.e.:
\begin{equation}
       \begin{array}{l l  c l}
    \xi : & \M & \rightarrow  & T \M   \\
             & x & \mapsto & \xi(x).
    \end{array}
\end{equation}
Notice that given this definition, $\xi(x) \in T_x \M~\forall~x\in \M$, so we also use the symbol $\xi$ (respectively $\eta$) to denote a tangent vector when there is no ambiguity.
Now, the Levi-Civita connection itself is defined as an operator
% \begin{equation}
%     \begin{array}{l l  c l}
%     \nabla : & T \M \times T \M & \rightarrow  & T \M   \\
%              & (\xi, \eta) & \mapsto & \nabla_\xi\eta
%     \end{array}
% \end{equation}
that generalizes the directional derivative of vectors fields to Riemannian manifolds, and associates to every couple of smooth vector fields $(\xi, \eta)$ on $\M$ a new vector field $\nabla_\xi\eta$ on $\M$.
Given a Riemannian manifold $\M$, the Levi-Civita connection is unique and defined by the \emph{Koszul} formula.
It should be noted that the Levi-Civita connection depends on the chosen Riemannian metric.
Using this object, a geodesic $\gamma: I \to \M$ with initial conditions $\gamma(0) = x$ and $\dot{\gamma}(0) = \xi$ is defined as a smooth curve having zero acceleration as defined by the Levi-Civita connection
\begin{equation}
	\nabla_{\dot{\gamma}(t)} \dot{\gamma}(t) = 0_{\gamma(t)}, \quad \forall t \in I
	\label{eq:zero_acc}
\end{equation}
where $\dot{\gamma}(t) = \frac{d}{dt} \gamma(t)$ and $0_{\gamma(t)}$ is the zero element of $T_{\gamma(t)}\M$.
%Then, the \emph{Riemannian exponential mapping} is introduced.
Let $\gamma$ be a geodesic defined on $[0, 1]$ with initial conditions $\gamma(0) = x$ and $\dot{\gamma}(0) = \xi$.
Then, the \emph{Riemannian exponential mapping} $\exp^\M_x: T_x\M \to \M$ at $x \in \M$ is defined as $\exp^\M_x(\xi) = \gamma(1)$.
For $x, y \in \M$, its inverse function, the \emph{Riemannian logarithm mapping}, is defined as $ \log^\M_x(y) = \argmin_{\xi \in T_x\M}  \norm{\xi}_x^2 \textup{ subject to } \exp^\M_x(\xi) = y$ with $\norm{\xi}_x^2 = \langle \xi, \xi \rangle_x^\M$.
Finally, the \emph{Riemannian distance} between two points $x, y \in \M$ is computed as $d_\M(x, y) = \lVert\log^\M_x(y)\rVert_x$. % = \left\langle \log^\M_x(y), \log^\M_x(y) \right\rangle_x^\M$.

\subsection{Description of the Riemannian manifold \texorpdfstring{$\Mmsg$}{Mmsg}}
\label{sec:IG_compound_Gaussian}

This subsection gives the Riemannian structure, induced by the FIM, of the parameter set $\Mmsg$.
% This Riemannian structure is very important for the rest of the paper.
% Indeed, we would like to minimize functions $h: \Mmsg \to \R$ which map mixtures of scaled Gaussian distributions to real values.
% To achieve this minimization, we provide a Riemannian based optimization Algorithm in the Section~\ref{sec:Riemannian_opt} which relies on this Riemannian structure of $\Mmsg$.
% Furthermore, numerical experiments in the Section~\ref{sec:num_exp} show that the speed of convergence of the minimization of functions $h$ depend on the provided Riemannian structure.
To specify the latter, we begin by defining the ambient space
\begin{equation}
	\label{eq:ambient_space}
	\Emsg = \R^p \times \R^{p \times p} \times \R^n.
\end{equation}
Therefore, the tangent space of $\Mmsg$ at $\theta$ is a subspace of the ambient space $\Emsg$
\begin{multline}
	\label{eq:tangent_space}
	T_\theta \Mmsg = \left\{ \xi = (\bXi_\bMu, \bXi_\bSigma, \bXi_\bt) \in \R^p \times \Sym \times \R^n:\right.\\ \left.\bXi_\bt^T \bt^{\odot -1} = 0 \right\} \, ,
\end{multline}
where $\Sym$ is the set of symmetric matrices and $.^{\odot -1}$ is the elementwise inverse operator.
To turn $\Mmsg$ into a Riemannian manifold, we must equip $\Mmsg$ with a Riemannian metric.
Many possibilities are available to us, however, a preferable one is the FIM~\cite{A16} derived in  Proposition~\ref{prop:FIM}.
Indeed, it is calculated using the NLL~\eqref{eq:NLL} and thus is associated with the statistical model~\eqref{eq:model}.

\begin{proposition}[Fisher Information Metric]
    \label{prop:FIM}
    Let $\theta \in \Mmsg$ and $\xi$, $\eta \in T_\theta \Mmsg$, the Fisher Information Metric at $\theta$ associated with the NLL~\eqref{eq:NLL} is
	\begin{multline*}
		\!\!\!\langle \xi, \eta \rangle_{\theta}^{\Mmsg} = \sum_{i=1}^n \bigg(\frac{1}{\tau_i}\bigg) \bXi_\bMu^T\bSigma^{-1}\bEta_\bMu 
		+ \frac{n}{2} \Tr \left( \bSigma^{-1} \bXi_\bSigma \bSigma^{-1}\bEta_\bSigma \right) \\
		+ \frac{p}{2}(\bXi_\bt \odot \bt^{\odot -1})^T(\bEta_\bt \odot \bt^{\odot -1})\, ,
	\end{multline*}
	where $\odot$ is the elementwise product operator.
\end{proposition}
\begin{proof}
	See supplementary material~\ref{supp:subsec:proof_FIM}.
\end{proof}

% \noindent
% Notably, the FIM computed in the Proposition~\ref{prop:FIM} is invariant under affine transformations.
% Given $\theta \in \Mmsg$, $\xi, \eta \in T_\theta\Mmsg$, $\bA \in \Gl$ (set of $p\times p$ invertible matrices) and $\bMu_0 \in \R^p$ we verify that
% \begin{equation}
% 	\langle \D\psi(\theta)[\xi], \D\psi(\theta)[\eta] \rangle_{\psi(\theta)}^{\Mmsg} = \langle \xi, \eta \rangle_{\theta}^{\Mmsg},
% \end{equation}
% where $\D \psi(\theta) [\xi]$ is the directional derivative of $\psi$ at $\theta$ in the direction $\xi$ with $\psi$ the affine transformation
% \begin{equation}
% 	\label{eq:aff_transformation}
% 	\psi(\theta) = \bigg(\bA \bMu + \bMu_0, \bA\bSigma\bA^T, \bt \bigg).
% \end{equation}
% This invariance property is of particular interest for problems that assume a mixing models such as blind signal separation~\cite{C98} or multi-spectral imaging~\cite{SS91}.
% Indeed, a mixing model assumes that the measured signal is a linear combination of a non-measurable source signal that is of interest for applications.
% and the FIM from the Proposition~\ref{prop:FIM} is invariant to this linear combination.

\noindent
% Restricted to elements of the tangent spaces $T_\theta \Mmsg$, the FIM from  Proposition~\ref{prop:FIM} defines a Riemannian metric on $\Mmsg$ which becomes a Riemannian manifold.
Then, the orthogonal projection according to the FIM from $\Emsg$ onto $T_\theta \Mmsg$ is given in Proposition~\ref{prop:orth_proj}.

\begin{proposition}[Orthogonal projection]
	\label{prop:orth_proj}
	The orthogonal projection associated with the FIM of Proposition~\ref{prop:FIM} from $\Emsg$ onto $T_\theta \Mmsg$ is
	\begin{equation*}
		P_\theta^\Mmsg(\xi) = \left(\bXi_\bMu, \sym(\bXi_\bSigma), \bXi_\bt - \frac{\bXi_\bt^T\bt^{\odot-1}}n \bt \right)\, ,
	\end{equation*}
	where $\sym(\bXi) = \frac12\left( \bXi + \bXi^T \right)$.
\end{proposition}
\begin{proof}
	See supplementary material~\ref{supp:subsec:proof_orth_proj}.
\end{proof}

\noindent
The orthogonal projection proves helpful to derive elements in tangent spaces such as the Riemannian gradient or the Levi-Civita connection.
The latter is given for the  manifold $\Mmsg$ in Proposition~\ref{prop:LVconnection}.

\begin{proposition}[Levi-Civita connection]
	\label{prop:LVconnection}
	Let $\theta \in \Mmsg$ and $\xi$, $\eta$ be smooth vector fields of $\Mmsg$, the Levi-Civita connection of $\Mmsg$ evaluated at $\theta$ is
	\begin{equation*}
		\nabla_\xi\eta = P_\theta^\Mmsg \left( \overbar{\nabla}_\xi\eta \right)\, ,
	\end{equation*}
	where
	\begin{multline*}
		\overbar{\nabla}_\xi\eta = \D\eta[\xi] +
			\Bigg (- \frac12 \Bigg[ \left(\frac{\bXi_\bt^T \bt^{\odot -2}}{\sum_{i=1}^n\frac{1}{\tau_i}} \bI_p + \bXi_\bSigma \bSigma^{-1} \right)\bEta_\bMu \\
				+ \left( \frac{\bEta_\bt^T\bt^{\odot -2}}{\sum_{i=1}^n\frac{1}{\tau_i}} \bI_p + \bEta_\bSigma \bSigma^{-1}\right)\bXi_\bMu \Bigg], \\
			       \frac1n \sum_{i=1}^n \bigg(\frac{1}{\tau_i}\bigg) \bEta_\bMu \bXi_\bMu^T - \bXi_\bSigma \bSigma^{-1}\bEta_\bSigma, \\
		       \frac1p \bXi_\bMu^T\bSigma^{-1}\bEta_\bMu  \1 - \bXi_\bt \odot \bEta_\bt \odot \bt^{\odot -1} \Bigg).
	\end{multline*}
\end{proposition}
\begin{proof}
	See supplementary material~\ref{supp:subsec:proof_LVconnection}.
\end{proof}

\noindent
As detailed in Subsection~\ref{subsec:intro_Riemannian_geometry} the Levi-Civita connection defines geodesics on a Riemannian manifold.
Indeed, for $I$ an open interval of $\R$, a geodesic $\gamma: I \to \Mmsg$ with initial position $\gamma(0) = \theta \in \Mmsg$ and initial velocity $\dot{\gamma}(0) = \xi \in T_\theta \Mmsg$ must respect
\begin{equation}
	\nabla_{\dot{\gamma}(t)} \dot{\gamma}(t) = 0_{\gamma(t)}, \quad \forall t \in I.
	\label{eq:zero_acc_Mmsg}
\end{equation}
However, an analytical solution of~\eqref{eq:zero_acc_Mmsg} remains unknown.
A retraction (approximation of the geodesic) can still be obtained (see Proposition~\ref{prop:retr}) which allows us to optimize functions on $\Mmsg$.
Moreover, the geodesic between two points $\theta_1$ and $\theta_2$ is unknown.
This implies that the geodesic distance is also unknown.
This is not surprising since the geodesic and the Riemannian distance between two Gaussian distributions with different locations are unknown~\cite{S84, CO91, PB16, TRZL19, CBGBRO22}.
To alleviate this problem, a divergence associated with the NC-MSG~\eqref{eq:distribution} is proposed in Section~\ref{sec:ML}.

\section{Riemannian optimization on \texorpdfstring{$\Mmsg$}{Mmsg}}
	\label{sec:Riemannian_opt}
	
\begin{algorithm}[t]
	\KwIn{Initialization $\theta^{(0)} \in \Mmsg$}
	\KwOut{Iterates $\theta^{(k)} \in \Mmsg$}
	\For{$k=0$ \textbf{to convergence}}{
		Compute a step-size $\alpha$ using Algorithm~\ref{algo:backtracking}\\
        Set $\theta^{(k+1)} \leftarrow R_{\theta^{(k)}}^\Mmsg \left(-\alpha \grad_\Mmsg h(\theta^{(k)})\right)$
	}
	\caption{Riemannian gradient descent on $\Mmsg$}
	\label{algo:steepest_IG}
\end{algorithm}

\begin{algorithm}[t]
    \KwIn{Current iterate $\theta^{(k)}\in \Mmsg$, and constants $\alpha \in ]0, t_\text{max}[$, $c \in ]0, 1[$ and $\varepsilon \in \RPos$}
	\KwOut{Step-size $\alpha$}
    Set $\theta(\alpha) \leftarrow R_{\theta^{(k)}}^\Mmsg \left(-\alpha\grad_\Mmsg h(\theta^{(k)})\right)$\\
	\While{$\! h(\theta^{(k)}) \! - \! h(\theta(\alpha)) \! < \! \varepsilon \alpha ||\grad_\Mmsg \!\! h(\theta^{(k)})||_{\theta^{(k)}}^2 \!\!\!$}{
        Set $\alpha \leftarrow c \,\alpha$\\
        Set $\theta(\alpha) \leftarrow R_{\theta^{(k)}}^\Mmsg \left(-\alpha\grad_\Mmsg h(\theta^{(k)})\right)$
	}
	\caption{Riemannian backtracking on $\Mmsg$}
	\label{algo:backtracking}
\end{algorithm}

The objective of this subsection is to propose tools to perform optimization on the Riemannian manifold $\Mmsg$.
Indeed, we aim to minimize smooth functions $h: \Mmsg \to \R$,
\begin{equation}
	\underset{\theta \in \Mmsg}{\textup{minimize}}\, \, h(\theta).
	\label{eq:min_h}
\end{equation}
An example of such a function is the NLL~\eqref{eq:NLL}.
As mentioned in Section~\ref{sec:Riemannian_geometry}, two additional cost functions are studied in Sections~\ref{sec:MLE} and~\ref{sec:ML}.
To realize~\eqref{eq:min_h}, we consider a Riemannian steepest gradient descent on $\Mmsg$.
Only the tools required for this algorithm are derived here.
For a detailed introduction to optimization on Riemannian manifolds, see~\cite{AMS08, B22}.
Two optimization tools are needed:
\emph{(i)} the Riemannian gradient of $h$,
\emph{(ii)} a retraction that maps tangent vectors from $T_\theta\Mmsg \,\, \forall \theta \in \Mmsg$ onto $\Mmsg$.
Once these are defined, the Riemannian steepest gradient descent retracts iteratively minus the gradient of $h$ times a step size onto the manifold.

We begin with the Riemannian gradient of $h$ at $\theta$.
For every $\theta \in \Mmsg$, it is defined through the Riemannian metric as the unique tangent vector in $T_\theta \Mmsg$ such that, $\forall \xi \in T_\theta\Mmsg$,
\begin{equation}
	\D h(\theta)[\xi] = \langle \grad_\Mmsg h(\theta), \xi \rangle_\theta^\Mmsg \, ,
\end{equation}
where $\D h(\theta) [\xi]$ is the directional derivative of $h$ at $\theta$ in the direction $\xi$.
In the case where for every $\theta \in \Mmsg$, there exists an open $U$ of $\Emsg$, with $\theta \in U$, and a differentiable function $\bar{h}: U \to \R$ such that $\bar{h}$ restricted to $\Mmsg$ is equal to $h$, this Riemannian gradient can be computed from the Euclidean gradient of $h$ at $\theta$.
In particular, this assumption is met by the different cost functions considered in the rest of the manuscript and the transformation of the Euclidean gradient into the Riemannian one is given in  Proposition~\ref{prop:grad}.
The latter is very convenient since this Euclidean gradient can be computed using automatic differentiation libraries such as Autograd~\cite{autograd} or JAX~\cite{jax}.
\begin{proposition}[Riemannian gradient]
	\label{prop:grad}
	Let $\theta \in \Mmsg$ and $h$ be a real-valued function defined on $\Mmsg$.
	The Riemannian gradient of $h$ at $\theta$ is
	\begin{multline*}
		\grad_\Mmsg h(\theta) =\\
		P_\theta^\Mmsg\Bigg(\left(\sum_{i=1}^n \frac{1}{\tau_i} \right)^{-1} \bSigma \G_\bMu, \frac2n \bSigma \G_\bSigma \bSigma, \frac2p \bt^{\odot 2} \odot \G_\bt \Bigg) \, ,
	\end{multline*}
	where $\grad h(\theta) = \left( \G_\bMu, \G_\bSigma, \G_\bt \right)$ is the Euclidean gradient of $h$ in $\R^p \times \R^{p\times p} \times \R^n$.
\end{proposition}
\begin{proof}
	See supplementary material~\ref{supp:subsec:proof_Riem_grad}.
\end{proof}

Then, it remains to define a retraction for every $\theta$ on $\Mmsg$.
A retraction $R_\theta^\Mmsg$ maps every $\xi \in T_\theta\Mmsg$ to a point $R_\theta^\Mmsg(\xi) \in \Mmsg$ and is such that $R_\theta^\Mmsg(\xi) = \theta + \xi + o(\norm{\xi})$.
Several retractions could be obtained from this definition.
Furthermore, it should be noted that a map respecting this definition is not necessarily related to the Riemannian metric of $\Mmsg$.
Thus, we choose to enforce an additional property: the desired retraction must have a zero initial acceleration, \emph{i.e.}
\begin{equation}
	\nabla_{\rdot(t)} \rdot(t) \evaltzero = 0\, ,
\end{equation}
where $\rdot(t) = \frac{d}{dt} R_\theta^\Mmsg(t\xi)$ and $\nabla$ is the Levi-Civita connection from Proposition~\ref{prop:LVconnection}.
Such a retraction is called a second-order retraction.
Furthermore, the property of zero initial acceleration is linked to the definition of the geodesic.
Indeed, a geodesic has a zero acceleration $\forall t$ along its path (see~\eqref{eq:zero_acc}) whereas here this condition is only needed at $t=0$.
By respecting this property, the retraction is associated with the Riemannian metric of  Proposition~\ref{prop:FIM} since the Levi-Civita connection is derived from this Riemannian metric.
Such a retraction is presented in  Proposition~\ref{prop:retr}.

\begin{proposition}[Second order retraction]
	\label{prop:retr}
	Let $\theta \in \Mmsg$ and $\xi \in T_\theta\Mmsg$.
	There exists $t_{max} > 0$ (specified in the Supplementary material) such that $\forall t \in [0, t_{max}[$, a second order retraction on $\Mmsg$ at $\theta$ is
	\begin{multline*}
		R_\theta^\Mmsg (t\xi) = \Bigg( \bMu + t\bXi_\bMu + \frac{t^2}{2} \left[\frac{\bXi_\bt^T \bt^{\odot -2}}{\sum_{i=1}^n\frac{1}{\tau_i}} \bI_p + \bXi_\bSigma \bSigma^{-1}\right]\bXi_\bMu ,\\
			\bSigma + t\bXi_\bSigma + \frac{t^2}{2} \left( \bXi_\bSigma \bSigma^{-1} \bXi_\bSigma -\frac1n \sum_{i=1}^n \left(\frac{1}{\tau_i}\right) \bXi_\bMu \bXi_\bMu^T \right),  \\
			N\left(\bt + t\bXi_\bt + \frac{t^2}{2} \left(  \bXi_\bt^{\odot 2} \odot \bt^{\odot -1} - \frac1p \bXi_\bMu^T\bSigma^{-1}\bXi_\bMu\1 \right) \right) \Bigg)\, ,
	\end{multline*}
	where $\forall \bx = (x_i)_{1\leq i \leq n} \in \RPosVec$, $N(\bx) = \left(\prod_{i=1}^n x_i \right)^{-\frac{1}{n}} \bx$.
\end{proposition}
\begin{proof}
	See supplementary material~\ref{supp:subsec:proof_retr}.
\end{proof}

\noindent
With this retraction and the Riemannian gradient from Proposition~\ref{prop:grad}, we have all the tools required to derive a Riemannian steepest descent.
The latter is presented in Algorithm~\ref{algo:steepest_IG}.
It should be noted that, in practice, the step size is chosen using a backtracking algorithm~\cite[Ch. 4]{B22}.
Given an initial step-size $\alpha \in ]0, t_\text{max}[$ with $t_\text{max}$ defined in Proposition~\ref{prop:retr}, the algorithm reduces $\alpha$ by a factor $c \in ]0, 1[$ until the Armijo–Goldstein condition is satisfied.
Given $\varepsilon \in \RPos$ (generally fixed at $10^{-4}$) and the tentative next iterate
\begin{equation}
    \theta(\alpha) = R_{\theta^{(k)}}^\Mmsg \left(-\alpha\grad_\Mmsg h(\theta^{(k)})\right),
\end{equation}
the Armijo–Goldstein condition writes 
\begin{equation}
    h(\theta^{(k)}) - h(\theta(\alpha)) \geq \varepsilon \alpha \norm{\grad_\Mmsg h(\theta^{(k)})}_{\theta^{(k)}}^2.
\end{equation}
This procedure is presented in Algorithm~\ref{algo:backtracking}.

\section{Parameter estimation of the non-centered mixture of scaled Gaussian distributions}
	\label{sec:MLE}
	\renewcommand{\arraystretch}{2}
\begin{table*}[t]
	\centering
	\begin{tabular}{ c|c|c } 
		\hline
		Name & $\Reg(\theta)$ & $\reg(x)$ \\
		\hline
		L1 penalty & $\norm{\left(\diag(\bt)\otimes\bSigma\right)^{-1} - \kappa^{-1} \bI_{n p}}_1 = \sum_{i,j} \left|\left(\tau_i \lambda_j\right)^{-1} - \kappa^{-1} \right|$ & $|x^{-1} - \kappa^{-1} |$ \\
		\hline
		L2 penalty & $\norm{\left(\diag(\bt)\otimes\bSigma\right)^{-1} - \kappa^{-1} \bI_{n p}}^2_2 = \sum_{i,j} \left(\left(\tau_i \lambda_j\right)^{-1} - \kappa^{-1} \right)^2$ & $(x^{-1} - \kappa^{-1})^2$ \\
		\hline
		% $d^2_\Spos(\diag(\bt)\otimes\bSigma, \kappa \bI_{n p}) = \sum_{i,j} \left( \log(\tau_i \lambda_j) - \log(\kappa) \right)^2$ & $\left(\log(x) - \log(\kappa)\right)^2$ & \makecell{Riemannian affine invariant distance}\\
		% \hline
		\makecell{Bures-Wasserstein\\squared distance} & $d^2_\textup{BW}\left((\diag(\bt)\otimes\bSigma)^{-1}, \kappa^{-1}\bI_{n p}\right) = \sum_{i,j} \left(\left(\tau_i \lambda_j\right)^{-\frac12} - \kappa^{-\frac12}\right)^2$ & $\left(x^{-\frac12} - \kappa^{-\frac12}\right)^2$ \\
		\hline
		\makecell{Gaussian\\KL divergence} & $\delta_\textup{KL}(\kappa\bI_{n p}, \diag(\bt)\otimes\bSigma) = \frac12 \left[\sum_{i,j} \big( \kappa\left(\tau_i \lambda_j\right)^{-1} + \log\left(\tau_i \lambda_j\right) \big) - np (1 + \log(\kappa)) \right]$
		& $\frac12 \left[ \kappa x^{-1} + \log(x) - (1+\log(\kappa)) \right] $ \\
		\hline
	\end{tabular}
	\caption{
		Examples of regularizations $\Reg$ respecting the Assumptions~\ref{assumption:reg_sum},~\ref{assumption:reg_lim} and~\ref{assumption:reg_div}.
		$\forall q \in \mathbb{N}^*$, $\norm{.}_q$ is the Schatten norm, \emph{i.e.} $\forall \bA \in \Sym$ $\norm{\bA}_q^q = \sum_i \left| \lambda_i \right|^q$ where $\lambda_i$ are the eigenvalues of $\bA$.
		The diagonal matrix with elements of $\bt$ is denoted $\diag(\bt)$.
		The Kronecker product between matrices is denoted $\otimes$.
	}
	\label{table:regularizations}
\end{table*}

In the previous subsection, tools to perform optimization on $\Mmsg$ have been developed.
In this subsection, the objective is to leverage these tools to estimate the parameters of an NC-MSG~\eqref{eq:distribution}.
In the following, we assume having $n \geq 1$ data points $\left\{ \bx_i \right\}_{i=1}^n \subset \R^p$.
The estimation of the parameters of the statistical model~\eqref{eq:distribution} is performed by maximizing the associated likelihood on $\Mmsg$:
\begin{equation}
	\underset{\theta \in \Mmsg}{\textup{minimize}}\, \, \NLL \left( \theta | \{\bx_i\}_{i=1}^n \right) \, ,
	\label{eq:min_NLL}
\end{equation}
where $\NLL$ is the NLL~\eqref{eq:NLL}.
However, the existence of a solution to this problem is not guaranteed.
To build intuition, we present a short example of a problematical case where $\bMu$ gets attracted by one data point $\bx_j$.
Let $k$ be the current iteration of a given optimizer of \eqref{eq:min_NLL}.
For $k \to +\infty$, if $\bMu^{(k)} \rightarrow \bx_j$ faster than $\tau_j^{(k)} \rightarrow  0$ and $\forall i \neq j$, $\tau_i^{(k)} \rightarrow +\infty$, then the quadratic form in $\NLL$~\eqref{eq:NLL} tends to zero, which is its minimum,
\begin{equation}
	\label{eq:bad_minimum_1}
	\sum_{i=1}^n \frac{(\bx_i - \bMu^{(k)})^T \left(\bSigma^{(k)}\right)^{-1} (\bx_i - \bMu^{(k)})}{\tau_i^{(k)}} \xrightarrow[k \to +\infty]{} 0\, .
\end{equation}
\sloppy Then, if an eigenvalue $\lambda^{(k)}$ of $\bSigma^{(k)}$ tends $0$ slower than the respective limits of $\bMu^{(k)}$, $\tau_i^{(k)}$ and $\tau_j^{(k)}$ and since $\sum_{i=1}^n \log \Det{\tau_i \bSigma} = n \log\Det{\bSigma}$, we obtain
\begin{equation}
	\label{eq:bad_minimum_2}
	\NLL ( \theta^{(k)} | \{\bx_i\}_{i=1}^n ) \xrightarrow[k \to +\infty]{} - \infty\, .
\end{equation}
Hence, depending on the data points $\left\{ \bx_i \right\}_{i=1}^n$, a solution of the problem~\eqref{eq:min_NLL} does not necessarily exist.

To overcome this issue, we propose a regularization approach to the NLL. Firstly, we prove that this allows the existence of a solution depending on some assumptions on the regularization term in \ref{subsec:Existence}. Some interpretations on the chosen regularization are next given in \ref{subsec:interpretations}, and finally, we study the robustness of the solution to rigid transformations in \ref{subsec:robustness}.

\subsection{Existence of solution with a regularized version of the NLL}
\label{subsec:Existence}

We present a regularized version of the NLL~\eqref{eq:NLL}:
\begin{equation}
	\label{eq:NLLReg}
	\NLLReg \left( \theta | \{\bx_i\}_{i=1}^n \right) = \NLL \left( \theta | \{\bx_i\}_{i=1}^n \right) + \beta \Reg(\theta) \, ,
\end{equation}
where $\beta \in \R_*^+$ and $\Reg: \Mmsg \to \R$ is a regularization.
Thus, the minimization problem~\eqref{eq:min_NLL} becomes
\begin{equation}
	\label{eq:min_NLLReg}
	\underset{\theta \in \Mmsg}{\textup{minimize}}\, \NLLReg \left( \theta | \{\bx_i\}_{i=1}^n \right).
\end{equation}
Though~\eqref{eq:min_NLLReg} is a generic formulation, we will focus on several proposals that ensure the existence of a solution.
% The problem~\eqref{eq:min_NLLReg} is general.
% Since the objective is to have the existence of a solution for the problem~\eqref{eq:min_NLLReg}, we focus on more specific regularizations.
The proposed approach is to rewrite $\Reg$ as a sum of regularizations $\reg$ on the eigenvalues of $\tau_i \bSigma$.
This rewriting is formalized in Assumption~\ref{assumption:reg_sum}.

\begin{assumption}
	The regularization $\Reg$ is a positive function that is a sum of regularizations on the eigenvalues of $\tau_i \bSigma$
	\begin{equation*}
		\Reg(\theta) = \sum_{i=1}^{n} \sum_{j=1}^{p}\reg(\tau_i\lambda_j) \, ,
		\label{eq:Reg}
	\end{equation*}
	where $\lambda_j \in \R_*^+$ are the eigenvalues of $\bSigma$ and $\reg: \R_*^+ \to \R$ is a continuous function.
	\label{assumption:reg_sum}
\end{assumption}
\noindent
In the following, we assume that $\Reg$ respects  Assumption~\ref{assumption:reg_sum}.
To prevent the eigenvalues of $\tau_i \bSigma$ from taking values that are too large or too small, a second Assumption is added.
Indeed, Assumption~\ref{assumption:reg_lim} states that the regularization of the $\log$ function by the penalty function $\reg$ goes to infinite when its argument goes to $0^+$ or $+\infty$.
This assumption is made so that if an eigenvalue of $\tau_i \bSigma$ tends to $0^+$ or $+\infty$ then $\NLLReg \to +\infty$.

\begin{assumption}
The following function admits the limit $\forall \beta \in \RPos$
	\begin{equation}
		\lim_{x \to \partial \RPos} \log(x) + \beta\reg(x) = +\infty \, , \label{eq:regularized_rk}
	\end{equation}
	with $\partial \RPos$ is a border of $\RPos$, \emph{i.e.} $0^+$ or $+\infty$.
	\label{assumption:reg_lim}
\end{assumption}

\noindent
Assumptions~\ref{assumption:reg_sum} and~\ref{assumption:reg_lim} are sufficient to solve the problem of existence stated earlier.
Indeed, when $\Reg$ respects these assumptions, Proposition~\ref{prop:existence_MLE} states that the problem~\eqref{eq:min_NLLReg} has a solution, \emph{i.e.} $\NLLReg$ admits a minimum in $\Mmsg$.
Finally,  Assumptions~\ref{assumption:reg_sum} and~\ref{assumption:reg_lim} are quite easy to meet in practice.
Indeed, several regularizations respecting these assumptions are proposed in  Table~\ref{table:regularizations}.

\begin{proposition}[Existence]
	\label{prop:existence_MLE}
	Under  Assumptions~\ref{assumption:reg_sum} and~\ref{assumption:reg_lim}, and $\forall \beta \in \RPos$, the regularized NLL
	\begin{equation*}
		%\NLLReg(\theta) = \sum_{i=1}^n \left[ \log \left|\tau_i \bSigma\right| + \frac{(\bx - \bMu)^T \bSigma^{-1} (\bx - \bMu)}{\tau_i} \right] + \beta \sum_{i=1}^{n} \sum_{j=1}^{p}\reg(\tau_i\lambda_j)
		\theta \mapsto \NLLReg \left( \theta | \{\bx_i\}_{i=1}^n \right) = \NLL \left( \theta | \{\bx_i\}_{i=1}^n \right) + \beta \Reg(\theta)\, ,
	\end{equation*}
	with $\NLL$ being the NLL~\eqref{eq:NLL}, admits a minimum in $\Mmsg$.
\end{proposition}

%Pour écrire la preuve, je m'appuie sur:
%\begin{itemize}
%    \item $[-\infty, +\infty]$ est un metric space : \href{https://math.stackexchange.com/questions/1901722/can-a-metric-be-evaluated-at-infinity}{url à cliquer}
%    \item $+\infty$ peut être un accumulation point :\href{https://math.stackexchange.com/questions/777031/how-can-infinity-be-an-accumulation-point}{url à cliquer}
%    \item dans un metric space $x^\star$ is an accumulation point of $x^{(k)}$ if there exists a sub-sequence $x^{(k_\ell)}$ such that $\lim_{k_\ell \to +\infty}x^{(k_\ell)} = x^\star$ : \href{https://math.stackexchange.com/questions/192463/accumulation-points-of-sequences-as-limits-of-subsequences}{url à cliquer}
%    \item decomposing a sequence with accumulation points into convergent sub-sequences: \href{https://math.stackexchange.com/questions/3273397/decomposing-a-sequence-with-countable-number-of-accumulation-points-into-converg}{url à cliquer}, \href{https://math.stackexchange.com/questions/1190595/limit-points-and-subsequences}{url à cliquer}
%    \item $x^{(k_\ell)}$ is a sub-sequence of $x^{(k)}$ then $x^{(k)} \to +\infty \iff x^{(k_\ell)} \to \infty$ : \href{https://math.stackexchange.com/questions/2459913/if-a-sequence-approaches-infinity-then-all-its-subsequences-approach-infinity}{url à cliquer}
%\end{itemize}

\begin{proof}
$\NLLReg$ is a continuous function on $\Mmsg$.
Hence, to prove the existence of a solution to the minimization problem~\eqref{eq:min_NLLReg}, it is enough to show that for sequences $\theta^{(k)} \to \partial \theta$, the boundary of $\Mpn$, we have that
\begin{equation}
   \label{eq:coercive_function}
    \lim_{k \to +\infty} \NLLReg(\theta^{(k)} | \{\bx_i\}_{i=1}^n) = +\infty\, .
\end{equation}
First, we handle the cases where $\norm{\bMu^{(k)}}_2 \centernot\longrightarrow +\infty$.
Since $\theta^{(k)} \to \partial \theta$, this means that, at least, one $\lambda^{(k)}_j \to \partial \RPos$ and/or one $\tau^{(k)}_i \to \partial \RPos$, with $\partial \RPos$ being the boundary of $\RPos$, \emph{i.e.} $\partial\RPos = \{0,  +\infty\}$.
Using the positivity of the quadratic form in the NLL~\eqref{eq:NLL}, we get the following inequality
\begin{equation}
    \label{eq:lower_bound_NLL}
    \NLL (\theta^{(k)} | \{\bx_i\}_{i=1}^n) \geq \sum_{i=1}^n \log\left|\tau^{(k)}_i \bSigma^{(k)} \right|\, .
\end{equation}
Hence, we get the resulting inequality on the regularized cost function
\begin{multline}
    \label{eq:lower_bound_NLLReg}
    \NLLReg(\theta^{(k)} | \{\bx_i\}_{i=1}^n) \geq\\ \sum_{i=1}^n \sum_{j=1}^p \left[ \log(\tau^{(k)}_i \lambda^{(k)}_j) + \beta \reg(\tau^{(k)}_i \lambda^{(k)}_j)\right]\, .
\end{multline}
 
%We remark that the lower bound~\eqref{eq:lower_bound_NLLReg} does not depend on $\bMu$.
Then, we give a sufficient condition to prove~\eqref{eq:coercive_function} when $\bSigma^{(k)} \to \partial \Spos$ and/or $\bt^{(k)} \to \partial \SRPosVec$, the boundaries of $\partial \Spos$ and $\partial \SRPosVec$ respectively.
To give this sufficient condition, we first recall Assumption~\ref{assumption:reg_sum}, $\forall \beta \in \RPos$
\begin{equation*}
    \lim_{x \to \partial \RPos} \log(x) + \beta r(x) = +\infty\, .
\end{equation*}
Thus, to prove~\eqref{eq:coercive_function}, a sufficient condition, when $\bSigma^{(k)} \to \partial \Spos$ and/or $\bt^{(k)} \to \partial \SRPosVec$ is that there exists at least one term $\tau^{(k)}_i \lambda^{(k)}_j$ such that
\begin{equation}
    \label{eq:sufficient_condition}
    \tau^{(k)}_i \lambda^{(k)}_j \to \partial \RPos\, .
\end{equation}
Since $\bSigma^{(k)} \to \partial \Spos$ and/or $\bt^{(k)} \to \partial \SRPosVec$, there exists at least one $\lambda^{(k)}_j \to \partial \RPos$ and/or one $\tau^{(k)}_i \to \partial \RPos$.

The condition~\eqref{eq:sufficient_condition} is of course met in the four following cases
\begin{equation*}
    \lambda^{(k)}_j \to 0^+ \textup{  and/or  } \tau^{(k)}_i \to 0^+,
\end{equation*}
\begin{equation*}
    \lambda^{(k)}_j \to +\infty \textup{  and/or  } \tau^{(k)}_i \to +\infty\, ,
\end{equation*}
\begin{equation*}
    \lambda^{(k)}_j \to 0^+ \textup{  and  } \tau^{(k)}_i \to +\infty \textup{  such that  } \tau^{(k)}_i \lambda^{(k)}_j \to \partial \RPos\, ,
\end{equation*}
\begin{equation*}
    \lambda^{(k)}_j \to +\infty \textup{  and  } \tau^{(k)}_i \to 0^+ \textup{  such that  } \tau^{(k)}_i \lambda^{(k)}_j \to \partial \RPos\, .
\end{equation*}
Finally, we treat the case where $\forall l \in \{1, \dotsb, n\}$, $\lambda^{(k)}_l \to \partial \RPos$ and $\tau^{(k)}_i \to \partial \RPos$ such that $\tau^{(k)}_i \lambda^{(k)}_l \centernot\longrightarrow \partial \RPos$.
Since $\prod_{m=1}^n \tau^{(k)}_m = 1$, there exists at least one $\tau_q$, with $q \neq i$, such that
\begin{equation}
    \label{eq:existence_taulambda_infty}
    \tau^{(k)}_q \lambda^{(k)}_j \to \partial \RPos.
\end{equation}
Hence, the condition~\eqref{eq:coercive_function} is met.

Before going further, we define the two following functions:
\begin{equation*}
    g_\Spos(\bSigma^{(k)}) = \norm{\log(\bSigma^{(k)})}_F^2 = \sum_{j=1}^p \log(\lambda_j^{(k)})^2 \, ,
\end{equation*}
and
\begin{equation*}
    g_\RPosVec(\bt^{(k)}) = \norm{\log(\bt^{(k)})}_2^2 =  \sum_{i=1}^n \log(\tau_i^{(k)})^2 \, .
\end{equation*}
It should be noted that $\sup (g_\Spos(\bSigma^{(k)})_{k=0}^{+\infty}) = +\infty$ if and only if there exists $j$ such that $(\lambda_j^{(k)})_{k=0}^{+\infty}$ has $\partial \RPos$ as accumulation point.
Similarly, $\sup(g_\RPosVec(\bt^{(k)})_{k=0}^{+\infty}) = +\infty$ if and only if there exists $i$ such that $(\tau_i^{(k)})_{k=0}^{+\infty}$ has $\partial \RPos$ as accumulation point.

Second, we consider the cases where $\norm{\bMu^{(k)}}_2 \to +\infty$, $\sup(g_\Spos(\bSigma^{(k)})_{k=0}^{+\infty}) < +\infty$ and $\sup(g_\RPosVec(\bt^{(k)})_{k=0}^{+\infty}) < +\infty$.
In this case, there exists $\lambda_\text{min}, \lambda_\text{max}>0$ and $\tau_\text{min}, \tau_\text{max}>0$ such that for all $k$, $\lambda_\text{min}\bI_p \preceq \bSigma^{(k)} \preceq \lambda_\text{max}\bI_p$ and $\tau_\text{min}\1 \leq \bt^{(k)} \leq \tau_\text{max}\1$.
Indeed, otherwise there would exist $j,i$ such that $\partial \RPos$ is an accumulation point of $(\lambda_j^{(k)})_{k=0}^{+\infty}$ and/or $(\tau_i^{(k)})_{k=0}^{+\infty}$.
Using these inequalities and the positivity of the regularization $\Reg$, we get that
\begin{equation*}
   \NLLReg(\theta^{(k)} | \{\bx_i\}_{i=1}^n) \geq \sum_{i=1}^n \frac{\norm{\bx_i - \bMu^{(k)}}^2}{\lambda_\text{max} \tau_\text{max}} + \text{const} \, ,
\end{equation*}
where the constant is independent from $\bMu^{(k)}$.
Thus, we have
\begin{equation*}
    \lim_{k \to +\infty} \NLLReg(\theta^{(k)} | \{\bx_i\}_{i=1}^n) = +\infty\, .
\end{equation*}
 
Third, it remains to check the cases where $\norm{\bMu^{(k)}} \to +\infty$, $\sup(g_\Spos(\bSigma^{(k)})_{k=0}^{+\infty}) = +\infty$ and/or $\sup(g_\RPosVec(\bt^{(k)})_{k=0}^{+\infty}) = +\infty$.
Thus, as said previously, there exists at least one $j$ and/or one $i$ such that $(\lambda_j^{(k)})_{k=0}^{+\infty}$ and/or $(\tau_i^{(k)})_{k=0}^{+\infty}$ has/have $\partial \RPos$ as accumulation point.
For each each of those $j,i$, we extract sub-sequences from $(\theta^{(k)})_{k=0}^{+\infty}$ whose limits in $\lambda_j$ and/or $\tau_i$ are these problematic accumulation points.
Then, we construct a partition of $\mathbb{N}$ with the indices corresponding to the elements of these sub-sequences and the indices of the remaining elements of the initial sequence $(\theta^{(k)})_{k=0}^{+\infty}$.
Let $\mathcal{I}$ be such partition of $\mathbb{N}$.
If $\mathcal{I}$ has a finite number of elements and if for every $(k_\ell)_{\ell=0}^{+\infty}\in\mathcal{I}$ we have
\begin{equation}
    \label{eq:coercive_function_subsequences}
    \lim_{\ell \to +\infty} \NLLReg(\theta^{(k_\ell)} | \{\bx_i\}_{i=1}^n) = +\infty\, ,
\end{equation}
then we get~\eqref{eq:coercive_function}.

% If  $\sup(g_\Spos(\bSigma^{(k)}))_{k=0}^{+\infty} = +\infty$ then there exists $j$ such that $(\lambda_j^{(k)})_{k=0}^{+\infty}$ has $\partial \RPos$ as accumulation point.
To do so, in the space $\RPosExt = [0, +\infty]$ equipped with the metric $d(x, y) = |\arctan(x) - \arctan(y)|$, given a $(\lambda_j^{(k)})_{k=0}^{+\infty}$ that has $L \in \partial \RPos$ as accumulation point, one can extract a sub-sequence of indices $(k_\ell)_{\ell=0}^{+\infty}$ such that $\lambda_j^{(k_\ell)} \to L$ and for $(k_m)_{m= 0}^{+\infty} = \mathbb{N} \setminus (k_\ell)_{\ell=0}^{+\infty}$,  $(\lambda_j^{(k_m)})_{m = 0}^{+\infty}$ does not have $L$ as accumulation point.
The same process is repeated iteratively from the remaining indices $(k_m)_{m=0}^{+\infty}$ for all $j$ such that $(\lambda_j^{(k_m)})_{m = 0}^{+\infty}$ still has an $L \in \partial \RPos$ as accumulation point.
It finishes when the sequence associated with the remaining elements of the original sequence $(\lambda_j^{(k)})_{k=0}^{+\infty}$ has no accumulation points in $\partial \RPos$.
Lets denotes $(k_q)_{q=0}^{+\infty}$ the remaining indices.
Then, the same process is also performed on $(\tau_i^{(k_q)})_{q=0}^{+\infty}$ if $\sup(g_\RPosVec(\bt^{(k_q)})_{q=0}^{+\infty}) = +\infty$.
% This way, we extract one sub-sequence $(k_\ell)_{\ell=0}^{+\infty}$ per accumulation point in $\partial \RPos$ of $(\lambda_j^{(k)})_{k=0}^{+\infty}$ and $(\tau_i^{(k)})_{k=0}^{+\infty}$ for all $j,i$.
All the obtained sequences of indices $(k_\ell)_{\ell=0}^{+\infty}$ along with the remaining elements of the original indices form a partition of $\mathbb{N}$. % without changing the corresponding limits in $\bSigma$ and $\bt$.
% Indeed, the intersection of two of these distinct sequences of indices has a finite number of elements or we would have a common sub-sequence with two distinct limits (otherwise the two sequences of indices can be merged).
% Hence, we redistribute these finite intersections until having a partition.
Due to its construction, this partition has at most $\card(\partial\RPos)^{p+n}+1 = 2^{p+n}+1$ elements.
Furthermore, we point out that, since $\norm{\bMu^{(k)}} \to +\infty$, we have that for every sub-sequence $(\bMu^{(k_\ell)})_{\ell=0}^{+\infty}$,  $\norm{\bMu^{(k_\ell)}} \to +\infty$.
Thus, for every $(k_\ell)_{\ell=0}^{+\infty} \in \mathcal{I}$, we have
\begin{itemize}
    \item either $\norm{\bMu^{(k_\ell)}} \to +\infty$, $\sup(g_\Spos(\bSigma^{(k_\ell)})_{\ell=0}^{+\infty}) < +\infty$ and $\sup(g_\RPosVec(\bt^{(k_\ell)})_{\ell=0}^{+\infty}) < +\infty$,
    \item or $\norm{\bMu^{(k_\ell)}} \to +\infty$ and there exists $i$ and/or $j$ such that $\lambda_j^{(k_\ell)} \to \partial\RPos$  and/or $\tau_i^{(k_\ell)} \to \partial\RPos$.
\end{itemize}
The former case has already been treated earlier.
For the latter case, one can reuse the arguments between~\eqref{eq:lower_bound_NLL} and~\eqref{eq:existence_taulambda_infty} to prove~\eqref{eq:coercive_function_subsequences}.
Indeed,~\eqref{eq:lower_bound_NLL} discards the quadratic form in $\bMu^{(k_\ell)}$ and hence the equations between~\eqref{eq:lower_bound_NLL} and~\eqref{eq:existence_taulambda_infty} hold.
Thus, the condition~\eqref{eq:coercive_function} is met.

\end{proof}

\subsection{Interpretations of the regularization term}
\label{subsec:interpretations}

So far, the regularization penalty has been chosen to guarantee the existence of a solution to the problem~\eqref{eq:min_NLLReg} without having specific insights on its impact on the estimate.
Therefore, this section thus discusses the interpretations of various classes of penalties and their related shrinkage effect.

A Bayesian interpretation of the considered penalties $\mathcal{R}_\kappa$ requires first discussing the case where it is decoupled in terms of $\{\tau_i\}_{i=1}^n$ and $\{\lambda_j\}_{j=1}^p$, i.e., when it can be expressed as
\begin{equation}
    \mathcal{R}_\kappa(\theta) 
    = 
    p \sum_{i=1}^n r^\tau_\kappa (\tau_i)
    +
    n \sum_{j=1}^p r^\lambda_\kappa (\lambda_j) \, .
\end{equation}
In such cases, 
\begin{itemize}
    \item[$\bullet$] 
    $r^\tau_\kappa$ can be linked to a pdf on $\tau$, denoted $f_\tau$.
    Assuming that
    $r^\lambda_\kappa(t)=0$ the optimization problem
    relates to the maximum a posteriori estimation of the Compound Gaussian model
    $\bx \sim \mathcal{N}(\mathbf{\bMu, \tau \bSigma})$ with
    $\tau \sim f_\tau$ \cite{OTKP12,M76}.
    Such a procedure is not often put into practice
    because it is generally possible (and preferable)
    to study the resulting pdf for the observations $\bx$:
    \begin{equation}
        f_{CG} (\bx) \propto   \int 
        f_G( \bx | \bMu, \tau \bSigma ) \, 
        f_\tau(\tau) \,  {\rm d} \tau \, ,
    \end{equation}
    whose MLE estimator appears as a special case of $M$-estimators of location and scatter, and is tractable with a fixed point algorithm \cite{OTKP12,M76}.
    \item[$\bullet$] 
    The penalty $r^\lambda_\kappa$ could also
    be interpreted as a pdf on the eigenvalues of $\bSigma$.
    This approach is less often studied from the Bayesian point of view because it does not have a clear interpretation
    of the distribution of the resulting $\bSigma$.
    Still, such penalties were leveraged to ensure
    existence of solutions of regularized $M$-estimators
    when $n<p$, e.g, in \cite{BOP19,YT20,W12}.
\end{itemize}
When additional prior information is available 
(power constraints that bound the eigenvalues, a rough estimate of the textures pdf, etc.) a Bayesian approach can be practically leveraged to select the form of the regularization penalty and the regularization parameters $\kappa$ and $\beta$.

In the general case of Assumption 1, i.e., where $\mathcal{R}_\kappa$ is possibly not decoupled, a Bayesian interpretation of $\mathcal{R}_\kappa$ is not as apparent.
Still, we can show that when the penalty can be interpreted as a divergence, it allows for explaining its effect on the estimate.
First, we recall the definition of a divergence:
\begin{definition}[Divergence]
	A divergence on a set $E$ is a function $\delta(.,.): E \times E \rightarrow \R$ satisfying, $\forall x, y \in E$:
	\begin{enumerate}
		\item $\delta(x, y) \geq 0$, %(non-negativity)
		\item $\delta(x, y) = 0 \iff x = y$\,.
	\end{enumerate}
	\label{def:div}
\end{definition}
\noindent
We can then state the following assumption, which is notably verified for all regularization examples given in Table 1:
\begin{assumption}
	The regularization $\Reg$ can be written as
	\begin{equation*}
		\Reg(\theta) = \delta_\Spos \left(\diag(\bt)\otimes\bSigma, \kappa \bI_{n p} \right)\, ,
	\end{equation*}
	where $\delta_\Spos$ is a divergence on the set $\Spos$ and $\kappa \in \RPos$. %between $\diag(\bt)\otimes\bSigma$ and $\kappa \bI_{n p}$ :
	\label{assumption:reg_div}
\end{assumption}
\noindent
This assumption allows us to state the following proposition:
\begin{proposition}[Minima of $\Reg$]
	\label{prop:minima_reg}
	Under the Assumption~\ref{assumption:reg_div}, the set of minima in $\Mmsg$ of the regularization $\Reg$ is
	\begin{equation*}
		\{ \theta = \left( \bMu, \kappa \bI_p, \1 \right) : \bMu \in \R^p \}\, .
	\end{equation*}
\end{proposition}
\begin{proof}
	The objective of this proof is to solve
	\begin{equation*}
		\underset{\theta \in \Mmsg}{\textup{minimize}}\, \,\Reg(\theta)\, .
	\end{equation*}
	Using Assumption~\ref{assumption:reg_div}, we know that $\Reg(\theta) \geq 0$ and $\Reg(\theta) = 0 \iff \diag(\bt) \otimes \bSigma = \kappa \bI_{n p}$.
	Thus, the minimum of $\Reg$ is $0$ and is reached at $\diag(\bt) \otimes \bSigma = \kappa \bI_{n p}$, $\forall \bMu \in \R^p$.
	This implies that the minimum satisfies the following system of equations
	\begin{equation*}
		\tau_i \, \lambda_j = \kappa \quad \forall i, j\, .
	\end{equation*}
	Hence, we deduce that $\tau_1 = \dotsm= \tau_n$.
	Using the constraint $\prod_{i=1}^n \tau_i = 1$, we get that $\tau_1 = \dotsm= \tau_n = 1$.
	Thus, $\lambda_1 = \dotsm = \lambda_p = \kappa$.
	This means that
	\begin{equation*}
		\left\{(\bMu, \kappa \bI_{p}, \1): \bMu \in \R^p\right\} = \argmin_{\theta \in \Mmsg} \Reg(\theta)\, ,
	\end{equation*}
	which characterizes Proposition~\ref{prop:minima_reg}.
\end{proof}

\noindent
Thus, under Assumption \ref{assumption:reg_div}, the minimum of~\eqref{eq:min_NLLReg} tends to $\left(\frac1n \sum_{i=1}^n \bx_i, \kappa \bI_p, \1 \right)$ as $\beta \to +\infty$.
This corresponds to the MLE of a Gaussian distribution with a covariance matrix $\kappa \bI$.
Thus, the hyperparameter $\beta$ makes the trade-off between an NC-MSG~\eqref{eq:distribution} and a white Gaussian distribution.
Hence, one can set in practice the hyperparameter $\kappa$ as $\kappa = \frac{1}{p}\Tr( \frac{1}{n} \bX\bX^T) =  \frac{1}{np} \sum_{i=1}^n \norm{\bx_i}_2^2$, meaning that the eigenvalues will be shrunk towards their empirical mean.
The effect of the regularization then echoes to existing shrinkage of $M$-estimators that have the same action
\cite{W12, OT14, PCQ14,SBP14, OPP21}.

To conclude, Assumption~\ref{assumption:reg_lim} and Proposition~\ref{prop:existence_MLE} provide the conditions that ensure the existence of a solution of the regularized MLE for any $\kappa>0$, whether $\mathcal{R}_\kappa$ is decoupled (with a Bayesian interpretation), interpretable as a divergence (following Assumption~\ref{assumption:reg_div} and Proposition~\ref{prop:minima_reg}), or not.
This class of regularization penalties thus allows going beyond the Bayesian estimation framework.
In practice, we mostly consider minimizing the estimation bias induced by the penalty and set $\beta$ close to $0$.
 For other tasks such as estimates used in classification, we resort to cross-validation procedures to select $\beta$ (see example in Figure~\ref{fig:num_exp:tuning_beta}).

\subsection{Robustness to rigid transformations}
\label{subsec:robustness}

We finish this section with a remark on estimating the parameter $\theta$ when data undergo a rigid transformation.
First of all, we define the set of orthogonal matrices
\begin{equation}
	\Op = \left\{ \bQ \in \R^{p\times p} : \bQ^T\bQ = \bI_p \right\}.
\end{equation}
Then, given $\bQ \in \Op$ and $\bMu_0 \in \R^p$, the rigid transformation $\psi$ of a set of data $\{ \bx_i \}_{i=1}^n$ is defined as
\begin{equation}
	\label{eq:rigid_transfo}
	\psi \left( \left\{ \bx_i \right\}_{i=1}^n \right) = \left\{ \bQ^T \bx_i + \bMu_0 \right\}_{i=1}^n.
\end{equation}
These rigid transformations define isometries on $\R^p$ since
\begin{equation}
	\norm{\psi\left(\bx_i\right) - \psi\left(\bx_j\right)}_2 = \norm{\bx_i - \bx_j}_2\, ,
\end{equation}
$\forall \bx_i, \bx_j \in \R^p$.
These are important in machine learning problems since they transform data without changing distances.
An important property of the regularized NLL~\eqref{eq:NLLReg} is that the estimated textures of the model are invariant under rigid transformations of the data; see Proposition~\ref{prop:minima_transformed_data}.
% for machine learning problems since transformations can happen between the training and the testing phases.
This is interesting since having parameters invariant to these transformations can improve performances when transformations happen between the training and the test sets for a given supervised problem.
Numerical experiments in  Section~\ref{sec:num_exp} leverage this property and show robust performances when data undergo a rigid transformation during the testing phase.

\begin{proposition}[Minima of $\NLLReg$ and rigid transformations]
	\label{prop:minima_transformed_data}
	Let $\Reg$ be a regularization satisfying  Assumption~\ref{assumption:reg_sum}, and $\theta^\star=(\bMu, \bSigma, \bt)$ be a minimum of the regularized NLL~\eqref{eq:min_NLLReg} computed on data $\left\{ \bx_i \right\}_{i=1}^n$,~\emph{i.e.} %$\theta \mapsto \NLLReg \left( \theta | \{\bx_i\}_{i=1}^n \right)$ defined in the equation~\eqref{eq:min_NLLReg}.
	\begin{equation*}
		\theta^\star \in \argmin_{\theta \in \Mmsg} \NLLReg \left( \theta | \{\bx_i\}_{i=1}^n \right),
	\end{equation*}
	\sloppy then, given $\bQ \in \Op$ and $\bMu_0 \in \R^p$, a minimum of the regularized NLL computed on the transformed data $\psi \left( \left\{ \bx_i \right\}_{i=1}^n \right) = \left\{ \bQ^T \bx_i + \bMu_0 \right\}_{i=1}^n$ is $\phi(\theta^\star) = \left( \bQ^T \bMu + \bMu_0, \bQ^T\bSigma \bQ, \bt \right)$, \emph{i.e.}
	\begin{equation*}
		\phi(\theta^\star) \in \argmin_{\theta \in \Mmsg} \NLLReg \left( \theta | \psi\left(\left\{ \bx_i \right\}_{i=1}^n\right) \right)\, .
	\end{equation*}
\end{proposition}
\begin{proof}
	First of all, given $\bQ \in \Op$ and $\bMu_0 \in \R^p$, one can check that 
	\begin{equation*}
		\NLL \left( \phi(\theta) | \psi\left(\left\{\bx_i\right\}_{i=1}^n\right) \right) = \NLL \left( \theta |\left\{\bx_i\right\}_{i=1}^n \right)\, ,
	\end{equation*}
	where $\NLL$ is the NLL defined in~\eqref{eq:NLL}, $\theta = (\bMu, \bSigma, \bt)$, $\phi(\theta) = (\bQ^T \bMu + \bMu_0, \bQ^T \bSigma \bQ, \bt)$ and $\psi$ is defined in equation~\eqref{eq:rigid_transfo}.
	Then, $\Reg$ satisfies Assumption~\ref{assumption:reg_sum} and thus only depends on the eigenvalues of the matrices $\tau_i \bSigma$.
	This implies that $\Reg ( \phi(\theta) ) = \Reg( \theta )$ and hence we get that
	\begin{equation*}
		\NLLReg \left( \phi(\theta) | \psi\left(\left\{\bx_i\right\}_{i=1}^n\right) \right) = \NLLReg \left( \theta | \left\{\bx_i\right\}_{i=1}^n \right)\, .
	\end{equation*}
	 This implies that if $
		\theta^\star \in \argmin_{\theta \in \Mmsg} \NLLReg \left( \theta | \{\bx_i\}_{i=1}^n \right)$, 
	then $\phi(\theta^\star) \in \argmin_{\theta \in \Mmsg} \NLLReg \left( \theta | \psi\left(\left\{\bx_i\right\}_{i=1}^n\right) \right)$, 
	which concludes the proof.
\end{proof}

\section{KL divergence and Riemannian center of mass}
	\label{sec:ML}
	In the previous section, we proposed to optimize the regularized NLL~\eqref{eq:min_NLLReg} of the NC-MSG~\eqref{eq:distribution}.
Once these parameters are estimated, they can be used as features for Riemannian classification/clustering algorithms~\cite{BBCJ12, TPM07, TPM08, FOP13}.
To do this classification/clustering, two tools are presented in this section.
Firstly, since no closed-form formula of the Riemannian distance on $\Mmsg$ is known, a divergence between pairs of parameters is defined.
The proposed one is the KL divergence between two NC-MSGs~\eqref{eq:distribution}.
It benefits from a simple closed-form formula presented in Subsection~\ref{subsec:KL}.
Secondly, simple classification algorithms, such as \emph{K-means} or the \emph{Nearest centroïd classifier}, rely on an algorithm to average parameters.
Thus, an algorithm to compute centers of mass of estimated parameters $\theta$ must be defined.
This center of mass is defined using the KL divergence and is presented in  Subsection~\ref{subsec:center_of_mass}.
Its computation is realized with  Algorithm~\ref{algo:steepest_IG}.

\subsection{KL divergence}
\label{subsec:KL}

% As explained in the introduction of this Section, a divergence is needed to implement classification algorithms on $\Mmsg$ such as \emph{K-means} or the \emph{Nearest centroïd classifier}.
Classification/clustering algorithms, such as \emph{K-means} or the \emph{Nearest centroïd classifier}, rely on a divergence between points.
Thus, it remains to define a divergence on $\Mmsg$.
The latter must be related to the NC-MSG~\eqref{eq:distribution}.
Indeed, the objective is to classify its parameters $\theta$.
In the context of measuring proximities between distributions admitting probability density functions, a classical divergence is the KL.
The latter measures the similarity between two probability density functions.
Definition~\ref{def:KL} gives the general formula of the KL divergence.
\begin{definition}[KL divergence]
	\label{def:KL}
	Given two probability density functions $p$ and $q$ defined on the sample space $\mathcal{X}$, the KL divergence is
	\begin{equation*}
		\delta_\textup{KL} (p, q) = \int_{\mathcal{X}} p(x) \log \left( \frac{p(x)}{q(x)} \right) dx \, .
	\end{equation*}
\end{definition}

\noindent
Applied to NC-MSGs, the KL divergence is derived from the Gaussian one and is presented in Proposition~\ref{prop:div}.
It benefits from a simple closed-form formula and therefore is of practical interest.
\begin{proposition}[KL divergence]
	\label{prop:div}
	\sloppy Given the r.v. $x = \left(\bx_1, \dotsc, \bx_n\right)$ and two NC-MSGs of probability density functions $p_{\theta_1}(x) = \prod_{i=1}^n f\left(\bx_i | (\bMu_1, \bSigma_1, \tau_{1,i})\right)$ and $p_{\theta_2}(x) = \prod_{i=1}^n f\left(\bx_i | (\bMu_2, \bSigma_2, \tau_{2,i})\right)$, the KL divergence is
	\begin{multline*}
		\delta_\textup{KL}(\theta_1, \theta_2) = \frac12 \Bigg( \sum_{i=1}^n \frac{\tau_{1, i}}{\tau_{2, i}}\Tr\left(\bSigma_2^{-1} \bSigma_1 \right) + \\ \sum_{i=1}^n \frac{1}{\tau_{2, i}} \Delta\bMu^T \bSigma_2^{-1} \Delta\bMu + n\log\left( \frac{\Det{\bSigma_2}}{\Det{\bSigma_1}}\right) - np \Bigg) \, ,
	\end{multline*}
	with $\Delta \bMu = \bMu_2 - \bMu_1$.
\end{proposition}
\begin{proof}
    The r.v. $x = \left(\bx_1, \dotsc, \bx_n\right)$ can be vectorized into $\bx = [\bx_1^T, \dotsc, \bx_n^T]^T \in \R^{np}$ which follows a multivariate Gaussian distribution of location the concatenation of the locations of $\bx_1, \dotsc, \bx_n$ and of block-diagonal covariance matrix whose elements are the covariance matrices of $\bx_1, \dotsc, \bx_n$.
    Thus, the KL divergence between the probability density functions $p_{\theta_1}$ and $p_{\theta_2}$ is the KL divergence between two multivariate Gaussian distributions whose covariance matrices are block diagonal. 
    Using the KL divergence between Gaussian distributions and the constraint $\prod_{i=1}^n \tau_{1,i} = \prod_{i=1}^n \tau_{2,i} = 1$, we get the desired formula.
\end{proof}

\noindent
Finally, this KL divergence is non-symmetrical.
% We choose to symmetrize it and we define its symmetrization  as
We rely on the classical symmetrization to define the proposed divergence $\delta_{\Mmsg}: \Mmsg \times \Mmsg \to \R$,
\begin{equation}
	\delta_{\Mmsg} (\theta_1, \theta_2) = \frac12 \left( \delta_\textup{KL}(\theta_1, \theta_2) + \delta_\textup{KL}(\theta_2, \theta_1) \right)\, .
	\label{eq:divergence}
\end{equation}

\subsection{Center of mass computation}
\label{subsec:center_of_mass}

To implement simple machine learning algorithms such as \emph{K-means} or the \emph{Nearest centroïd classifier} on $\Mmsg$, it remains to define an averaging algorithm.
To do so, we leverage a classical definition of centers of mass which are minimizers of variances~\cite{K77, M05}.
Given a set of parameters $\{\theta_i\}_{i=1}^M$, its center of mass on $\Mmsg$ is defined as the solution of
\begin{equation}
	\underset{\theta \in \Mmsg}{\textup{minimize}}\, \, \frac1M \sum_{i=1}^M \delta_{\Mmsg}(\theta, \theta_i) \, ,
	\label{eq:min_center_of_mass}
\end{equation}
where $\delta_\Mmsg$ is the symmetrized KL divergence from Equation~\eqref{eq:divergence}.
To realize~\eqref{eq:min_center_of_mass}, Algorithm~\ref{algo:steepest_IG} can be employed.

\section{Numerical experiments}
	\label{sec:num_exp}
	The objective of this section is to show the practical interests of the tools developed in the previous sections.
More precisely, this section presents numerical experiments and is divided into two parts.

First, the subsection~\ref{sec:num_exp:simu} studies the performance of Algorithm~\ref{algo:steepest_IG}, in terms of speed of convergence on the cost functions~\eqref{eq:min_NLLReg} and~\eqref{eq:min_center_of_mass} and in terms of estimation error on the cost function~\eqref{eq:min_NLL}.
Both studies are done through simulations.
Algorithm~\ref{algo:steepest_IG} is shown to be fast.
Indeed, it requires from $5$ to $30$ times fewer iterations to minimize costs functions~\eqref{eq:min_NLLReg} and~\eqref{eq:min_center_of_mass} compared to other sophisticated optimization algorithms.
This demonstrates the interest in the choice of the FIM to develop Riemannian optimization algorithms.
Also, Algorithm~\ref{algo:steepest_IG} applied to the cost function~\eqref{eq:min_NLL} gives lower estimation errors than other classical estimators such as the Tyler joint mean-scatter one and the Gaussian ones.

Second, an application on the crop classification dataset \emph{Breizhcrops}~\cite{breizhcrops2020} is presented in Subsection~\ref{sec:num_exp:appli}.
This dataset consists of $600\ 000$ time series to be classified into $9$ classes.
The application implements a \emph{Nearest centroïd classsifier} on $\Mmsg$ using the divergence~\eqref{eq:divergence} and the Riemannian center of mass~\eqref{eq:min_center_of_mass}.
Three results ensue.
First, the proposed algorithms can be used on large-scale datasets.
Second, the proposed regularization in  Section~\ref{sec:MLE} plays an important role in classification.
Third, considering an NC-MSG~\eqref{eq:distribution} is interesting for time series especially when data undergo a rigid transformation~\eqref{eq:rigid_transfo}.

Python code implementing the different experiments can be found at \url{https://github.com/antoinecollas/optim_compound}.

\subsection{Simulation}
\label{sec:num_exp:simu}

% Numerical experiments on simulated data are presented in this subsection.
% First of all, a fast convergence of the Algorithm~\ref{algo:steepest_IG} on the minimization of the regularized NLL~\eqref{eq:min_NLLReg} and on the estimation of the center of mass~\eqref{eq:min_center_of_mass} is shown in the subsection~\ref{sec:num_exp:speed}.
% Then, the subsection~\ref{sec:num_exp:perf_estim} presents the low error of estimation made by the Algorithm~\ref{algo:steepest_IG} compared to other classical estimators.
% This estimation is realized on the parameters of the NC-MSG~\eqref{eq:distribution}  through the minimization of the NLL~\eqref{eq:min_NLL}.

In this simulation setting, we set the parameters $\theta = \left( \bMu, \bSigma, \bt \right) \in \Mmsg$ as follows.
First, each component of $\bMu$ is sampled from a univariate Gaussian distribution $\N(0, 1)$.
Second, $\bSigma$ is generated using its eigendecomposition $\bSigma = \U \bL \U^T$.
$\U \in \Op$ is drawn from the uniform distribution on $\Op$~\cite{M07} using the module ``scipy.stats" from the Scipy library~\cite{Scipy}.
Then, the elements on the diagonal of the diagonal matrix $\bL$ are drawn from a $\chi_1^2$ distribution.
Third, the $\tau_i$ are drawn from a $\Gamma(\nu, 1/\nu)$ distribution with $\nu$ a parameter to be chosen.
The smaller the $\nu$, the greater the variance.
In order to respect the constraint $\prod_{i=1}^n \tau_i = 1$, the vector $\bt$ is normalized.
% We point out that the following numerical experiments could be realized with other generation procedures for the parameter $\theta$.

% \subsubsection{Speed of convergence}
% \label{sec:num_exp:speed}

The speed of convergence of Algorithm~\ref{algo:steepest_IG} is studied on two cost functions: the regularized NLL~\eqref{eq:min_NLLReg} and the cost function~\eqref{eq:min_center_of_mass} to compute the center of mass associated to the KL divergence of Proposition~\ref{prop:div}.

\begin{figure*}
	\centering
	\includegraphics[width=\linewidth]{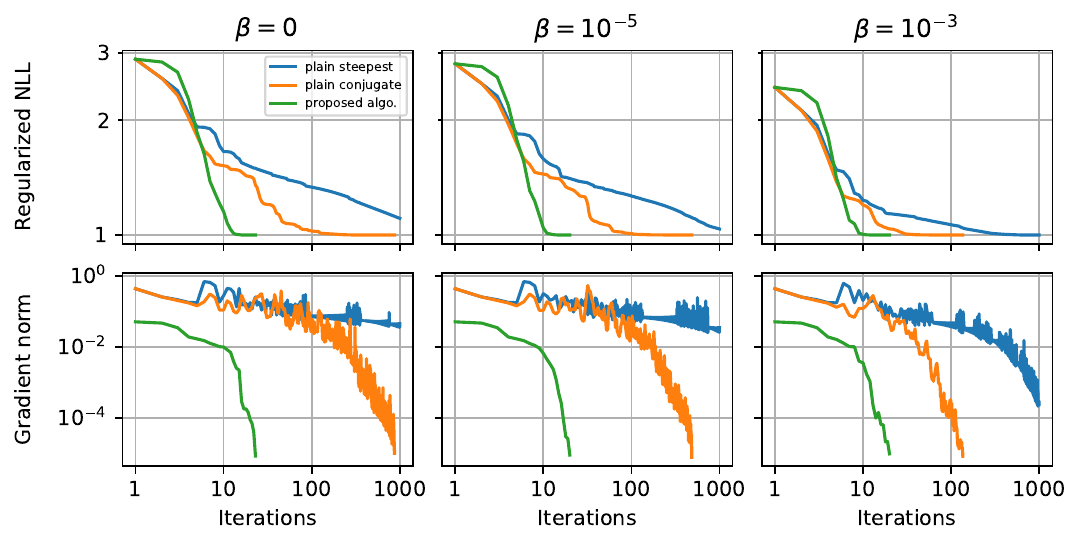}
	\caption{
		Regularized NLL~\eqref{eq:min_NLLReg} and its gradient norm versus the iterations of three estimation algorithms.
		The chosen regularization is the L2 penalty (see Table~\ref{table:regularizations}) and three different regularization intensities $\beta$ are considered: $0$ in the left column, $10^{-5}$ in the middle one, and $10^{-3}$ in the right one.
		Each estimation is performed on $n=150$ samples in $\R^{10}$ sampled from an NC-MSG.
		The regularized NLL are normalized so that their minimum value is $1$.
	}
	\label{fig:num_exp:speed_NLL_iterations}
\end{figure*}

\begin{figure*}
	\centering
	\includegraphics[width=\linewidth]{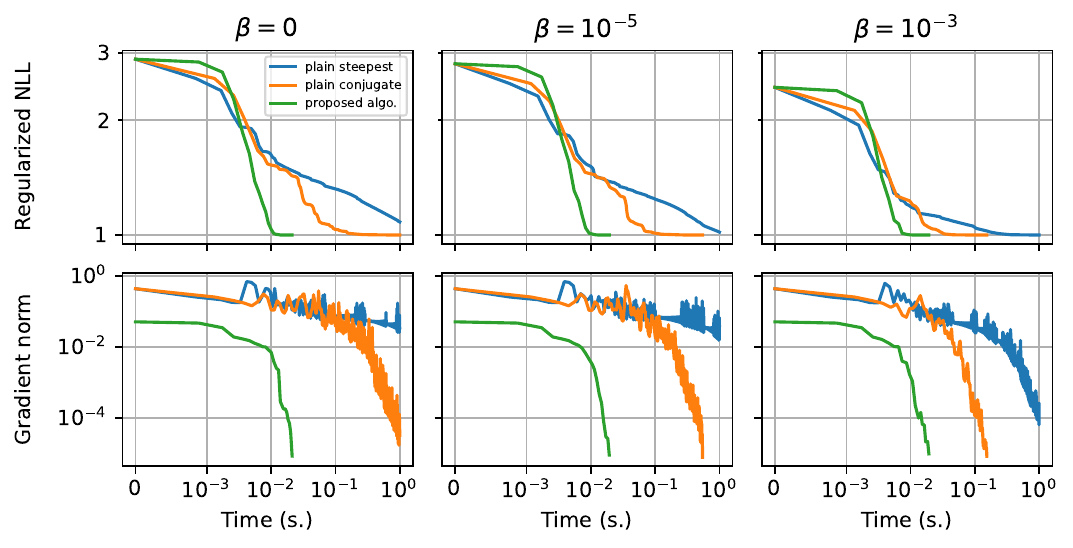}
	\caption{
		Regularized NLL~\eqref{eq:min_NLLReg} and its gradient norm versus the computation time of three estimation algorithms.
		The chosen regularization is the L2 penalty (see Table~\ref{table:regularizations}) and three different regularization intensities $\beta$ are considered: $0$ in the left column, $10^{-5}$ in the middle one, and $10^{-3}$ in the right one.
		Each estimation is performed on $n=150$ samples in $\R^{10}$ sampled from an NC-MSG.
		The regularized NLL are normalized so that their minimum value is $1$.
    }
	\label{fig:num_exp:speed_NLL_time}
\end{figure*}

\sloppy We begin with the minimization of the regularized NLL~\eqref{eq:min_NLLReg}.
$n=150$ data $\bx_i \in \R^{10}$ are drawn from a NC-MSG, \emph{i.e.} $\bx_i \sim \N(\bMu, \tau_i \bSigma)$.
The parameter $\theta = (\bMu, \bSigma, \bt)$ of this distribution is generated as explained in the introduction of this subsection with $\nu=1$.
Different parameters $\beta$ in~\eqref{eq:min_NLLReg} are considered: $\beta \in \{0, 10^{-5}, 10^{-3}\}$.
The chosen regularization is the L2 penalty from Table~\ref{table:regularizations}.
When $\beta = 0$ the NLL is the plain one, \emph{i.e.} it is not regularized.
We point out that, in this setup, the optimization goes well although the existence of a solution to this problem is not proven.
When $\beta > 0$ a solution to the minimization problem exists from Proposition~\ref{prop:existence_MLE}.
The minimization is performed with three different algorithms.
\begin{itemize}
	\item The plain conjugate gradient presented in~\cite{CBBRGO21}. It is a Riemannian conjugate gradient descent that uses a sum of three independent Riemannian metrics associated with the three parameters $\bMu$, $\bSigma$, and $\bt$. Thus, the corresponding Riemannian geometry is easier to derive but is not linked to the NC-MSG.
	\item The plain steepest descent. It is similar to the plain conjugate gradient. Still, it only uses the gradient as a direction of descent (and not a linear combination with the direction of descent of the previous step).
	\item The Algorithm~\ref{algo:steepest_IG} that leverages the information geometry of the NC-MSG presented in Section~\ref{sec:IG_compound_Gaussian}. 
\end{itemize}
The results of this experiment are presented in Figures~\ref{fig:num_exp:speed_NLL_iterations} and~\ref{fig:num_exp:speed_NLL_time} in terms of iterations and computation time respectively.
We observe that Algorithm~\ref{algo:steepest_IG} is much faster than the two others regardless of the $\beta$ parameter.
Indeed, in the case $\beta \in \{0, 10^{-5}\}$, the Algorithm~\ref{algo:steepest_IG} is at least $100$ times faster than the plain steepest descent and $10$ times faster than the plain conjugate gradient.
In the case of $\beta = 10^{-3}$, Algorithm~\ref{algo:steepest_IG} is at least $20$ times faster than the plain steepest descent and $3$ times faster than the plain conjugate gradient.
Furthermore, we observe these results are valid either in the number of iterations or in computation time.
Indeed, the three considered algorithms have iterations with similar computational costs in $\mathcal{O}(np^2 + p^3)$.
Thus, a reduction in the number of iterations results in a reduction in computation time.

\begin{figure*}
	\centering
	\includegraphics[width=\linewidth]{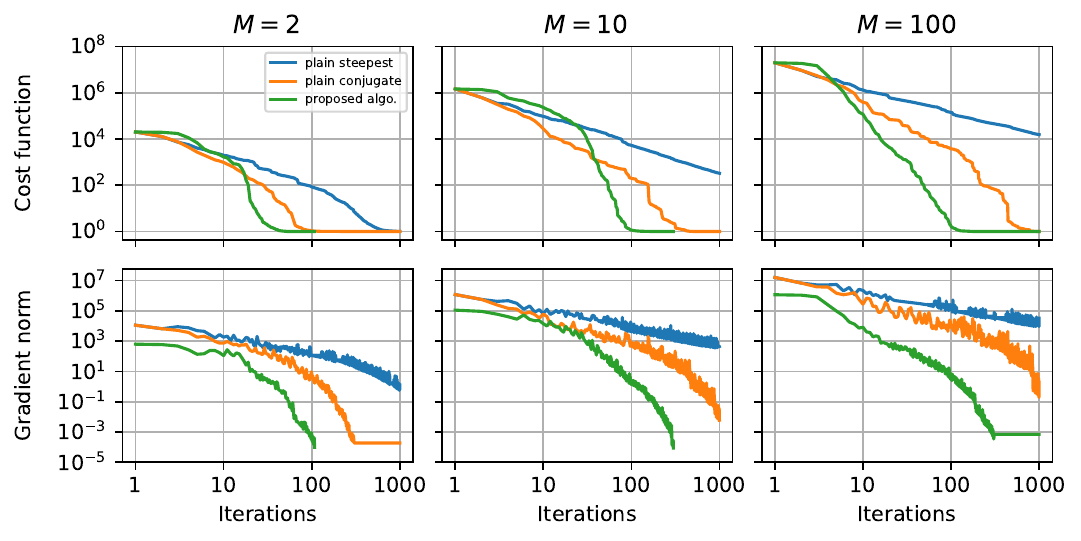}
	\caption{
		Cost function~\eqref{eq:min_center_of_mass} and its gradient norm versus the iterations of three estimation algorithms.
		The dimensions of the parameter space are $p=10$ and $n=150$.
		Three different numbers of points $M$ are considered: $2$ in the left column, $10$ in the middle one, and $100$ in the right one.
		The cost functions are normalized so that their minimum value is $1$.
	}
	\label{fig:num_exp:speed_cost_center_of_mass_iterations}
\end{figure*}

\begin{figure*}
	\centering
	\includegraphics[width=\linewidth]{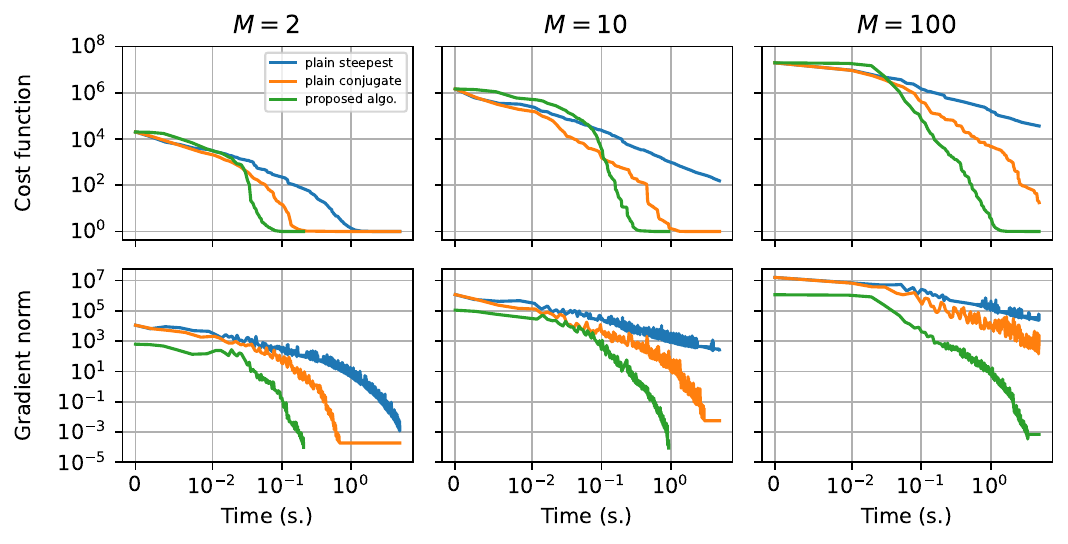}
	\caption{
		Cost function~\eqref{eq:min_center_of_mass} and its gradient norm versus the computation time of three estimation algorithms.
		The dimensions of the parameter space are $p=10$ and $n=150$.
		Three different numbers of points $M$ are considered: $2$ in the left column, $10$ in the middle one, and $100$ in the right one.
		The cost functions are normalized so that their minimum value is $1$.
    }
	\label{fig:num_exp:speed_cost_center_of_mass_time}
\end{figure*}

Then, a similar experiment is performed with the cost function~\eqref{eq:min_center_of_mass} to compute the center of mass.
% In this cost function, $\delta_{\M_{p,n}}(\theta, \theta_i) = \frac12 \left( \delta_\textup{KL}(\theta, \theta_i) + \delta_\textup{KL}(\theta_i, \theta) \right)$ is chosen to show the interest of the Algorithm~\ref{algo:steepest_IG}.
% Indeed, the associated minimization problem~\eqref{eq:min_center_of_mass} has no closed-form solution and the plain conjugate gradient and steepest descent are typically slow since they require several hundreds of iterations before reaching convergence.
$M \in \{2, 10, 100\}$ parameters $\theta$ are generated as described in the introduction of Subsection~\ref{sec:num_exp:simu} with $\nu = 1$.
The minimization is performed with the same optimization algorithms as previously: the plain steepest descent, the plain conjugate gradient, and Algorithm~\ref{algo:steepest_IG}.
The results of this experiment are presented in  Figures~\ref{fig:num_exp:speed_cost_center_of_mass_iterations} and~\ref{fig:num_exp:speed_cost_center_of_mass_time} in terms of iterations and computation time, respectively.

We observe that Algorithm~\ref{algo:steepest_IG} is much faster than the two others regardless of $M$.
Indeed, when $M=2$, Algorithm~\ref{algo:steepest_IG} converges in $40$ iterations whereas the plain conjugate gradient requires $300$ iterations and the plain steepest descent still has not converged after $1000$ iterations.
When $M \in \{10, 100\}$, Algorithm~\ref{algo:steepest_IG} converges in less than $60$ iterations which is $4$ times faster than the plain conjugate gradient.
It should be noted that the plain steepest descent has not converged after $1000$ iterations in the cases $M \in \{100, 1000\}$.
Once again, these results are valid either in the number of iterations or in computation time since the three considered algorithms have iterations with similar computational costs in $\mathcal{O}\left(M\left(n + p^3\right)\right)$.
Hence, reducing the number of iterations implies a reduction in computation time.

% \subsubsection{Estimation error}
% \label{sec:num_exp:perf_estim}

% This subsection studies the estimation error between estimated parameters and the true ones with numerical experiments on simulated data.
The estimation error made by Algorithm~\ref{algo:steepest_IG} applied on the NLL~\eqref{eq:NLL} is studied with numerical experiments on simulated data.
$n \in \llbracket 20, 1000 \rrbracket$ data $\bx_i$ are sampled from the NC-MSG~\eqref{eq:distribution}.
The parameter $\theta = (\bMu, \bSigma, \bt)$ of this distribution is generated as presented in the introduction of Subsection~\ref{sec:num_exp:simu} with $\nu = 0.1$ to have heterogeneous textures $\tau_i$.
The considered estimators for this numerical experiment are the following:
\begin{itemize}
	\item Gaussian estimators: the sample mean $\bMuG = \frac{1}{n} \sum_{i=1}^n \bx_i$ and the SCM $\bSigmaG = \frac{1}{n} \sum_{i=1}^n (\bx_i - \bMuG) \left(\bx_i - \bMuG \right)^T$.
	\item Tyler's joint location-scatter matrix estimator~\cite{T87} denoted $\bMuT$ and $\bSigmaT$.
	\item Tyler's $M$-estimator with location known~\cite{T87}. The sampled data $\bx_i$ are centered with the true location $\bMu$, and then $\bSigma$ is estimated. This estimator is denoted $\bSigmaTmu$.
	\item The proposed estimator denoted $\bMuIG$ and $\bSigmaIG$.  Algorithm~\ref{algo:steepest_IG} minimizes the NLL~\eqref{eq:NLL}.
    The initialization is the Gaussian maximum likelihood \emph{i.e.} $\theta_\textup{init} = \left(\bMuG, \bSigmaG, \mathbf{1}_n \right)$, where $\bMuG = \frac{1}{n} \sum_{i=1}^n \bx_i$, $\bSigmaG = \frac{1}{n} \sum_{i=1}^n \left(\bx_i - \bMuG\right) \left(\bx_i - \bMuG\right)^T$ and $\mathbf{1}_n = (1, \cdots, 1)^T$.
\end{itemize}

\newcommand\heightMSE{5cm}
\newcommand\scalelegendMSE{0.55}
\begin{figure}[t]
	\centering
    \includegraphics[width=0.85\linewidth]{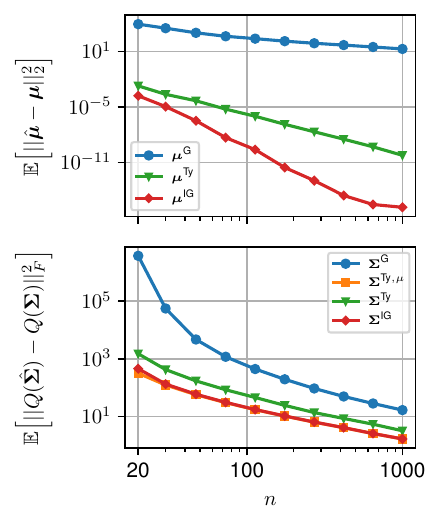}
	\caption{
		MSE over $2000$ simulated sets $\{\bx_i\}_{i=1}^n \subset \R^{10}$ versus the number samples $\bx_i$ for the considered estimators $\widehat{\bMu} \in \{\bMu^\textup{G}, \bMu^\textup{Ty}, \bMu^\textup{IG}\}$ and $\widehat{\bSigma} \in \{\bSigma^\textup{G}, \bSigma^{\textup{Ty}, \bMu}, \bSigma^\textup{Ty}, \bSigma^\textup{IG}\}$.
		The proposed estimators $\bMu^\textup{IG}$ and $\bSigma^\textup{IG}$ are computed as in~\eqref{eq:min_NLL} using Algorithm~\ref{algo:steepest_IG} and the $Q$ function normalizes scatter matrices, \emph{i.e.} $\forall \bSigma \in \Spos$, $Q(\bSigma) = \Det{\bSigma}^{-\frac{1}{p}}\bSigma$.
	}
	\label{fig:num_exp:error_estimation}
\end{figure}

The estimation errors are measured with the Mean Squared Errors (MSE).
These errors are computed as $\mathbb{E}[\norm{\hat{\bMu} - \bMu}_2^2]$ and $\mathbb{E}[\lVert Q(\hat{\bSigma}) - Q(\bSigma) \rVert_F^2]$, with $Q(\bSigma) = \Det{\bSigma}^{-\frac{1}{p}}\bSigma$, for the estimated location $\hat{\bMu}$ and the estimated scatter $\hat{\bSigma}$ respectively, with $2000$ Monte-Carlo.
The MSE on the location and the scatter versus the number of samples $\bx_i$ are plotted in Figure~\ref{fig:num_exp:error_estimation}.
First, we observe in both figures that the Gaussian estimators have a high MSE.
This shows the interest in considering robust estimators such as Tyler's joint location-scatter matrix estimator or the proposed one when the textures $\tau_i$ are heterogeneous.
Then, the proposed estimators realize a much lower MSE than Tyler's joint location-scatter estimator.
We can note that when enough samples are provided, the MSE on the location realized by the proposed estimator reaches the machine precision and is therefore negligible.
Finally, we compare the performance of the proposed estimator with Tyler's $M$-estimator for the scatter estimation.
Indeed, when the location is known, Tyler's $M$-estimator is the MLE of the NC-MSG~\eqref{eq:distribution}.
We observe that when enough samples are provided, the proposed estimator matches the MSE of Tyler's $M$-estimator.
Overall, this experimental subsection illustrates the good performance of the proposed estimator when data are sampled from a NC-MSG~\eqref{eq:distribution}.

\subsection{Application}
\label{sec:num_exp:appli}

\begin{figure}[t]
	\input{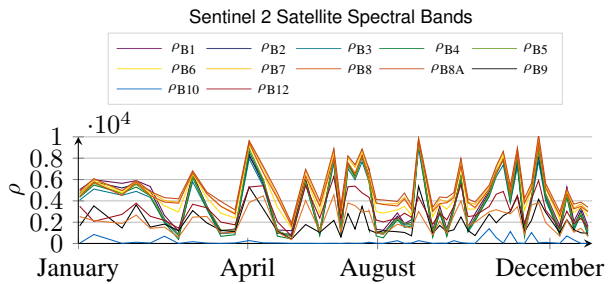}
	\examplemeadows
	\caption{
		Reflectances of a Sentinel-2 time series of meadows from the \emph{Breizhcrops} dataset.
		Figure courtesy~\cite{breizhcrops2020}.
	}
	\label{fig:num_exp:bzh}
\end{figure}

\newcommand\heightbeta{5cm}
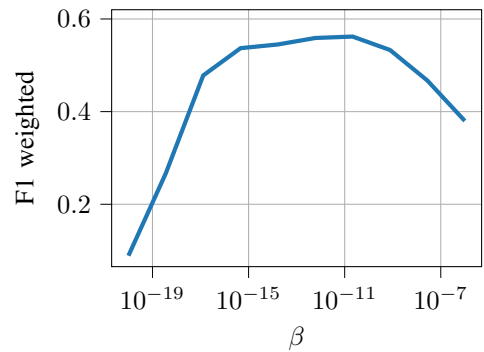
\begin{figure}[t]
	\centering
	\scalebox{1}{% This file was created with tikzplotlib v0.10.1.
\begin{tikzpicture}

\definecolor{darkgray176}{RGB}{176,176,176}
\definecolor{lightgray204}{RGB}{204,204,204}
\definecolor{steelblue31119180}{RGB}{31,119,180}

\begin{axis}[
height=\heightbeta,
width=1.3*\heightbeta,
legend cell align={left},
legend style={fill opacity=0.8, draw opacity=1, text opacity=1, draw=lightgray204},
log basis x={10},
tick align=outside,
tick pos=left,
xlabel={$\beta$},
ylabel={F1 weighted},
x grid style={darkgray176},
xmajorgrids,
xmin=1.99526231496888e-21, xmax=5.01187233627272e-06,
xmode=log,
xtick style={color=black},
xtick={1e-23,1e-19,1e-15,1e-11,1e-07,1e-03},
% xticklabels={
%   \(\displaystyle {10^{-23}}\),
%   \(\displaystyle {10^{-21}}\),
%   \(\displaystyle {10^{-19}}\),
%   \(\displaystyle {10^{-17}}\),
%   \(\displaystyle {10^{-15}}\),
%   \(\displaystyle {10^{-13}}\),
%   \(\displaystyle {10^{-11}}\),
%   \(\displaystyle {10^{-9}}\),
%   \(\displaystyle {10^{-7}}\),
%   \(\displaystyle {10^{-5}}\),
%   \(\displaystyle {10^{-3}}\)
% },
y grid style={darkgray176},
ymajorgrids,
ymin=0.06535, ymax=0.62,
ytick style={color=black},
]
\addplot [ultra thick, steelblue31119180]
table {%
1e-20 0.089
3.59381366380464e-19 0.267
1.29154966501488e-17 0.478
4.64158883361279e-16 0.537
1.66810053720006e-14 0.545
5.99484250318942e-13 0.559
2.15443469003189e-11 0.562
7.74263682681128e-10 0.533
2.78255940220713e-08 0.467
1e-06 0.380
};
% \addlegendentry{L2}
\end{axis}

\end{tikzpicture}}
	\caption{
	``F1 weighted" metric achieved by the proposed \emph{Nearest centroïd classifier} on the \emph{Breizhcrops} dataset versus the parameter of regularization $\beta$ in~\eqref{eq:NLLReg}.
		The chosen regularization is the L2 penalty from  Table~\ref{table:regularizations}.
	}
	\label{fig:num_exp:tuning_beta}
\end{figure}

\newcommand\heightappli{5cm}
\newcommand\scalelegendappli{0.65}
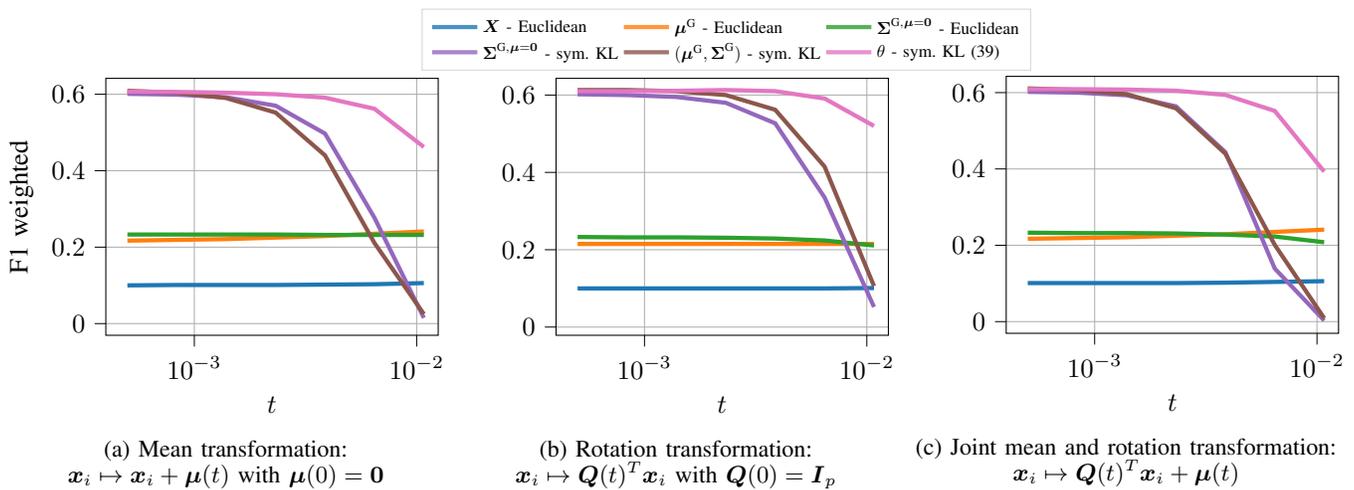
\begin{figure*}[t]
	\centering
	\captionsetup[subfigure]{justification=centering}
	\begin{subfigure}{0.33\linewidth}
		\centering
		% \resizebox{0.8\linewidth}{!}{\input{figures/classification/OA_vs_mean_transfo_.tex}}
		\scalebox{1}{% This file was created with tikzplotlib v0.10.1.
\begin{tikzpicture}

\definecolor{crimson2143940}{RGB}{214,39,40}
\definecolor{darkgray176}{RGB}{176,176,176}
\definecolor{darkorange25512714}{RGB}{255,127,14}
\definecolor{forestgreen4416044}{RGB}{44,160,44}
\definecolor{lightgray204}{RGB}{204,204,204}
\definecolor{mediumpurple148103189}{RGB}{148,103,189}
\definecolor{orchid227119194}{RGB}{227,119,194}
\definecolor{sienna1408675}{RGB}{140,86,75}
\definecolor{steelblue31119180}{RGB}{31,119,180}

\begin{axis}[
height=\heightappli,
width=1.2*\heightappli,
legend cell align={left},
legend style={
  legend style={nodes={scale=\scalelegendappli, transform shape}},
  legend columns=3,
  fill opacity=0,
  draw opacity=0.8,
  text opacity=1,
  at={(0.95, 1.03)},
  anchor=south west,
  draw=lightgray204,
},
log basis x={10},
tick align=outside,
tick pos=left,
x grid style={darkgray176},
xmajorgrids,
xlabel={$t$},
xmin=4e-04, xmax=1.25e-2,
xmode=log,
xtick style={color=black},
xtick={1e-07,1e-06,1e-05,0.0001,0.001,0.01,0.1},
xticklabels={
  \(\displaystyle {10^{-7}}\),
  \(\displaystyle {10^{-6}}\),
  \(\displaystyle {10^{-5}}\),
  \(\displaystyle {10^{-4}}\),
  \(\displaystyle {10^{-3}}\),
  \(\displaystyle {10^{-2}}\),
  \(\displaystyle {10^{-1}}\)
},
y grid style={darkgray176},
ymajorgrids,
ylabel={F1 weighted},
ymin=-0.03055, ymax=0.64155,
ytick style={color=black}
]
\addplot [ultra thick, steelblue31119180]
table {%
0.0005 0.1
0.00083405026860003 0.101
0.00139127970110356 0.101
0.00232079441680639 0.101
0.00387131841340563 0.102
0.00645774832507442 0.103
0.0107721734501594 0.106
% 0.0179690683190231 0.11
% 0.029974212515947 0.116
% 0.05 0.125
};
\addlegendentry{$\bX$ - Euclidean}
\addplot [ultra thick, darkorange25512714]
table {%
0.0005 0.217
0.00083405026860003 0.219
0.00139127970110356 0.221
0.00232079441680639 0.225
0.00387131841340563 0.229
0.00645774832507442 0.235
0.0107721734501594 0.241
% 0.0179690683190231 0.236
% 0.029974212515947 0.224
% 0.05 0.213
};
\addlegendentry{$\bMuG$ - Euclidean}
\addplot [ultra thick, forestgreen4416044]
table {%
0.0005 0.233
0.00083405026860003 0.233
0.00139127970110356 0.233
0.00232079441680639 0.233
0.00387131841340563 0.232
0.00645774832507442 0.232
0.0107721734501594 0.232
% 0.0179690683190231 0.23
% 0.029974212515947 0.227
% 0.05 0.203
};
\addlegendentry{$\bSigmaGcentered$ - Euclidean}
\addplot [ultra thick, mediumpurple148103189]
table {%
0.0005 0.601
0.00083405026860003 0.599
0.00139127970110356 0.592
0.00232079441680639 0.57
0.00387131841340563 0.497
0.00645774832507442 0.277
0.0107721734501594 0.015
% 0.0179690683190231 0.002
% 0.029974212515947 0.002
% 0.05 0.001
};
\addlegendentry{$\bSigmaGcentered$ - sym. KL}
\addplot [ultra thick, sienna1408675]
table {%
0.0005 0.609
0.00083405026860003 0.604
0.00139127970110356 0.59
0.00232079441680639 0.552
0.00387131841340563 0.44
0.00645774832507442 0.211
0.0107721734501594 0.025
% 0.0179690683190231 0.006
% 0.029974212515947 0.003
% 0.05 0.002
};
\addlegendentry{$(\bMuG, \bSigmaG)$ - sym. KL}
\addplot [ultra thick, orchid227119194]
table {%
0.0005 0.607
0.00083405026860003 0.606
0.00139127970110356 0.604
0.00232079441680639 0.6
0.00387131841340563 0.591
0.00645774832507442 0.562
0.0107721734501594 0.462
% 0.0179690683190231 0.147
% 0.029974212515947 0.01
% 0.05 0.002
};
% \addlegendentry{Location_covariance_texture_constrained_texture_div_alpha_sym_0_reg_L2_1e-11}
\addlegendentry{$\theta$ - sym. KL~\eqref{eq:divergence}}
\end{axis}

\end{tikzpicture}}
		\caption{
			Mean transformation:\\
			$\bx_i \mapsto \bx_i + \bMu(t)$
			with $\bMu(0) = \VEC{0}$ \\
			% \
		}
		% \label{subfig:num_exp:mean_transfo}
	\end{subfigure}%
	\begin{subfigure}{0.33\linewidth}
		\centering
		% \resizebox{0.8\linewidth}{!}{\input{figures/classification/OA_vs_rotation_transfo_.tex}}
		\scalebox{1}{% This file was created with tikzplotlib v0.10.1.
\begin{tikzpicture}

\definecolor{crimson2143940}{RGB}{214,39,40}
\definecolor{darkgray176}{RGB}{176,176,176}
\definecolor{darkorange25512714}{RGB}{255,127,14}
\definecolor{forestgreen4416044}{RGB}{44,160,44}
\definecolor{lightgray204}{RGB}{204,204,204}
\definecolor{mediumpurple148103189}{RGB}{148,103,189}
\definecolor{orchid227119194}{RGB}{227,119,194}
\definecolor{sienna1408675}{RGB}{140,86,75}
\definecolor{steelblue31119180}{RGB}{31,119,180}

\begin{axis}[
height=\heightappli,
width=1.2*\heightappli,
legend cell align={left},
legend style={
  legend style={nodes={scale=\scalelegendappli, transform shape}},
  legend columns=2,
  fill opacity=0,
  draw opacity=0.8,
  text opacity=1,
  at={(0, 1.03)},
  anchor=south west,
  draw=lightgray204,
},
log basis x={10},
tick align=outside,
tick pos=left,
x grid style={darkgray176},
xmajorgrids,
xlabel={$t$},
xmin=4e-04, xmax=1.25e-2,
xmode=log,
xmode=log,
xtick style={color=black},
xtick={1e-05,0.0001,0.001,0.01,0.1,1},
xticklabels={
  \(\displaystyle {10^{-5}}\),
  \(\displaystyle {10^{-4}}\),
  \(\displaystyle {10^{-3}}\),
  \(\displaystyle {10^{-2}}\),
  \(\displaystyle {10^{-1}}\),
  \(\displaystyle {10^{0}}\)
},
y grid style={darkgray176},
ymajorgrids,
ylabel={\phantom{F1 weighted}},
ymin=-0.0212, ymax=0.6432,
ytick style={color=black}
]
\addplot [ultra thick, steelblue31119180]
table {%
0.0005 0.1
0.00083405026860003 0.1
0.00139127970110356 0.1
0.00232079441680639 0.1
0.00387131841340563 0.1
0.00645774832507442 0.1
0.0107721734501594 0.101
% 0.0179690683190231 0.101
% 0.029974212515947 0.102
% 0.05 0.103
};
% \addlegendentry{$\bX$ - Euclidean}
\addplot [ultra thick, darkorange25512714]
table {%
0.0005 0.215
0.00083405026860003 0.215
0.00139127970110356 0.215
0.00232079441680639 0.215
0.00387131841340563 0.215
0.00645774832507442 0.215
0.0107721734501594 0.215
% 0.0179690683190231 0.215
% 0.029974212515947 0.215
% 0.05 0.214
};
% \addlegendentry{$\bMuG$ - Euclidean}
\addplot [ultra thick, forestgreen4416044]
table {%
0.0005 0.233
0.00083405026860003 0.232
0.00139127970110356 0.232
0.00232079441680639 0.231
0.00387131841340563 0.229
0.00645774832507442 0.224
0.0107721734501594 0.211
% 0.0179690683190231 0.179
% 0.029974212515947 0.115
% 0.05 0.06
};
% \addlegendentry{$\bSigmaGcentered$ - Euclidean}
\addplot [ultra thick, mediumpurple148103189]
table {%
0.0005 0.602
0.00083405026860003 0.6
0.00139127970110356 0.595
0.00232079441680639 0.58
0.00387131841340563 0.527
0.00645774832507442 0.335
0.0107721734501594 0.052
% 0.0179690683190231 0.01
% 0.029974212515947 0.009
% 0.05 0.005
};
% \addlegendentry{$\bSigmaGcentered$ - sym. KL}
\addplot [ultra thick, sienna1408675]
table {%
0.0005 0.613
0.00083405026860003 0.613
0.00139127970110356 0.61
0.00232079441680639 0.6
0.00387131841340563 0.562
0.00645774832507442 0.415
0.0107721734501594 0.106
% 0.0179690683190231 0.015
% 0.029974212515947 0.01
% 0.05 0.006
};
% \addlegendentry{$(\bMuG, \bSigmaG)$ - sym. KL}
\addplot [ultra thick, orchid227119194]
table {%
0.0005 0.61
0.00083405026860003 0.61
0.00139127970110356 0.611
0.00232079441680639 0.613
0.00387131841340563 0.61
0.00645774832507442 0.591
0.0107721734501594 0.52
% 0.0179690683190231 0.337
% 0.029974212515947 0.209
% 0.05 0.151
};
% \addlegendentry{$\theta$ - sym. KL~\eqref{eq:divergence}}
\end{axis}

\end{tikzpicture}}
		\caption{
			Rotation transformation:\\
			$\bx_i \mapsto \bQ(t)^T\bx_i$
			with $\bQ(0) = \bI_p$ \\
			% \
		}
		% \label{subfig:num_exp:rotation_transfo}
	\end{subfigure}%
	\begin{subfigure}{0.33\linewidth}
		\centering
		% \resizebox{0.8\linewidth}{!}{\input{figures/classification/OA_vs_mean_rotation_transfo_.tex}}
		\scalebox{1}{% This file was created with tikzplotlib v0.10.1.
\begin{tikzpicture}

\definecolor{crimson2143940}{RGB}{214,39,40}
\definecolor{darkgray176}{RGB}{176,176,176}
\definecolor{darkorange25512714}{RGB}{255,127,14}
\definecolor{forestgreen4416044}{RGB}{44,160,44}
\definecolor{lightgray204}{RGB}{204,204,204}
\definecolor{mediumpurple148103189}{RGB}{148,103,189}
\definecolor{orchid227119194}{RGB}{227,119,194}
\definecolor{sienna1408675}{RGB}{140,86,75}
\definecolor{steelblue31119180}{RGB}{31,119,180}

\begin{axis}[
height=\heightappli,
width=1.2*\heightappli,
legend cell align={left},
legend style={
  legend style={nodes={scale=\scalelegendappli, transform shape}},
  legend columns=2,
  fill opacity=0,
  draw opacity=0.8,
  text opacity=1,
  at={(0, 1.03)},
  anchor=south west,
  draw=lightgray204,
},
log basis x={10},
tick align=outside,
tick pos=left,
x grid style={darkgray176},
xmajorgrids,
xlabel={$t$},
xmin=4e-04, xmax=1.25e-2,
xmode=log,
xtick style={color=black},
xtick={1e-07,1e-06,1e-05,0.0001,0.001,0.01,0.1},
xticklabels={
  \(\displaystyle {10^{-7}}\),
  \(\displaystyle {10^{-6}}\),
  \(\displaystyle {10^{-5}}\),
  \(\displaystyle {10^{-4}}\),
  \(\displaystyle {10^{-3}}\),
  \(\displaystyle {10^{-2}}\),
  \(\displaystyle {10^{-1}}\)
},
y grid style={darkgray176},
ymajorgrids,
ylabel={\phantom{F1 weighted}},
ymin=-0.0306, ymax=0.6426,
ytick style={color=black}
]
\addplot [ultra thick, steelblue31119180]
table {%
0.0005 0.101
0.00083405026860003 0.101
0.00139127970110356 0.101
0.00232079441680639 0.101
0.00387131841340563 0.102
0.00645774832507442 0.104
0.0107721734501594 0.106
% 0.0179690683190231 0.11
% 0.029974212515947 0.116
% 0.05 0.125
};
% \addlegendentry{$\bX$ - Euclidean}
\addplot [ultra thick, darkorange25512714]
table {%
0.0005 0.217
0.00083405026860003 0.219
0.00139127970110356 0.221
0.00232079441680639 0.225
0.00387131841340563 0.229
0.00645774832507442 0.235
0.0107721734501594 0.241
% 0.0179690683190231 0.237
% 0.029974212515947 0.224
% 0.05 0.213
};
% \addlegendentry{$\bMuG$ - Euclidean}
\addplot [ultra thick, forestgreen4416044]
table {%
0.0005 0.233
0.00083405026860003 0.232
0.00139127970110356 0.232
0.00232079441680639 0.231
0.00387131841340563 0.228
0.00645774832507442 0.223
0.0107721734501594 0.208
% 0.0179690683190231 0.165
% 0.029974212515947 0.087
% 0.05 0.056
};
% \addlegendentry{$\bSigmaGcentered$ - Euclidean}
\addplot [ultra thick, mediumpurple148103189]
table {%
0.0005 0.602
0.00083405026860003 0.6
0.00139127970110356 0.593
0.00232079441680639 0.564
0.00387131841340563 0.444
0.00645774832507442 0.139
0.0107721734501594 0.003
% 0.0179690683190231 0.002
% 0.029974212515947 0.001
% 0.05 0.001
};
% \addlegendentry{$\bSigmaGcentered$ - sym. KL}
\addplot [ultra thick, sienna1408675]
table {%
0.0005 0.611
0.00083405026860003 0.607
0.00139127970110356 0.596
0.00232079441680639 0.559
0.00387131841340563 0.44
0.00645774832507442 0.201
0.0107721734501594 0.009
% 0.0179690683190231 0.002
% 0.029974212515947 0.001
% 0.05 0.001
};
% \addlegendentry{$(\bMuG, \bSigmaG)$ - sym. KL}
\addplot [ultra thick, orchid227119194]
table {%
0.0005 0.609
0.00083405026860003 0.609
0.00139127970110356 0.608
0.00232079441680639 0.605
0.00387131841340563 0.594
0.00645774832507442 0.552
0.0107721734501594 0.393
% 0.0179690683190231 0.097
% 0.029974212515947 0.013
% 0.05 0.005
};
% \addlegendentry{$\theta$ - sym. KL~\eqref{eq:divergence}}
\end{axis}

\end{tikzpicture}}
		\caption{
			Joint mean and rotation transformation:\\
			$\bx_i \mapsto \bQ(t)^T\bx_i + \bMu(t)$
			% with $\bMu(0) = \VEC{0}$ and $\bQ(0) = \bI_p$
		}
		% \label{subfig:num_exp:mean_rotation_transfo}
	\end{subfigure}%
	\caption{
		``F1 weighted" metric versus the parameter $t$ associated with three transformations applied to the test set of the \emph{Breizhcrops} dataset.
		The different \emph{Nearest centroïd classifiers} estimate the barycentres on the training data without transformations.
		Then, the classification is performed on the test set with three different transformations.
		For $t=0$, the test set is not transformed, and the larger $t$ is, the more the test set is transformed.
		Six different \emph{Nearest centroïd classifiers} are compared: each is a combination of an estimator, a divergence, and its associated center of mass computation.
		The proposed one is denoted ``$\theta$ - sym. KL".
		The latter uses the Equations~\eqref{eq:min_NLLReg},~\eqref{eq:divergence} and~\eqref{eq:min_center_of_mass} for the estimation, the divergence and the center of mass computation respectively.
		The regularization is the L2 penalty from  Table~\ref{table:regularizations} and $\beta$ is fixed at $10^{-11}$.
	}
	\label{fig:num_exp:rigid_transfo}
\end{figure*}

In the previous subsection, the different theoretical results derived in Sections from~\ref{sec:Riemannian_geometry} to~\ref{sec:ML} showed several interests in synthetic data.
We now focus on applying a \emph{Nearest centroïd classifier} on $\Mmsg$ to real data using the estimation framework developed in  Section~\ref{sec:MLE}, the divergence and the Riemannian center of mass from  Section~\ref{sec:ML} as well as the optimization framework from Section~\ref{sec:Riemannian_opt}.
This classifier is compared with several other \emph{Nearest centroïd classifiers} associated with different estimators and divergences.

To do so, we consider the dataset \emph{Breizhcrops}~\cite{breizhcrops2020}: a large-scale dataset of more than $600\ 000$ crop time series from the Sentinel-2 satellite to classify.
More specifically, for each crop $n = 45$ observations $\bx_i \in \R^p$ are measured over time.
Each $\bx_i$ contains reflectance measurements of $p = 13$ spectral bands.
Then, these measurements are concatenated into one batch $\bX_j = [\bx_1, \cdots, \bx_n] \in \R^{p\times n}$.
Hence, we get one matrix $\bX_j$ per crop and each one belongs to an unknown class $y \in \llbracket 1, K \rrbracket$.
These $K=9$ classes represent crop types such as nuts, barley, or wheat and are heavily imbalanced, \emph{i.e.} some classes are much more represented than others.
An example of a time series of meadows is presented in  Figure~\ref{fig:num_exp:bzh}.
We apply a single preprocessing step: all the data are centered using the global mean.
For simplicity, the matrix $\bX_j$ is noted $\bX$ in the following.

To classify these crops, we apply a \emph{Nearest centroïd classifier} on descriptors.
Indeed, the use of statistical descriptors is a classical procedure in machine learning as they are often more discriminative than raw data (see \emph{e.g.} \cite{BBCJ12, TPM07}).
Hence, this classification algorithm works in three steps.
\begin{enumerate}
	\item For each batch $\bX$, a descriptor is computed, \emph{e.g.} the parameter $\theta \in \Mmsg$ from the minimization of the regularized NLL~\eqref{eq:NLLReg}.
	\item Then, on the training set, the center of mass of the descriptors of each class is computed. This center of mass is always computed by minimizing the variance associated with a divergence between descriptors. For example, the center of mass on $\Mmsg$ is computed as in~\eqref{eq:min_center_of_mass}.
	\item Finally, on the test set, each descriptor is labeled with the class of the nearest center of mass with respect to the chosen divergence.
\end{enumerate}
% Thus, we get a classification of the batches $\bX$.

Six~\emph{Nearest centroïd classifiers} are considered, and they are grouped according to the divergence they use: the Euclidean distance, the symmetrized KL divergence between Gaussian distributions, or the symmetrized KL divergence~\eqref{eq:divergence} between NC-MSGs.
For each divergence, several \emph{Nearest centroïd classifiers} are derived using several estimators.
These estimators correspond to different assumptions on the data.

Three \emph{Nearest centroïd classifiers} rely on the Euclidean distance between matrices. Given two matrices $\bA$ and $\bB$ of the same size, the Euclidean distance is $d(\bA, \bB) = \norm{\bA - \bB}_F$.
The center of mass of a given set $\left\{\bA_i\right\}_{i=1}^M$ is the arithmetic mean $ \frac{1}{M} \sum_{i=1}^M \bA_i$ which is the solution of $\minimize_{\bY} \frac{1}{M} \sum_{i=1}^M \norm{\bY - \bA_i}_F^2$.
From this geometry, three \emph{Nearest centroïd classifiers} are derived using three estimators: the batch itself $\bX$, the sample mean $\bMuG$ and $\bSigmaGcentered = \frac{1}{n} \sum_{i=1}^n \bx_i \bx_i^T$.
The last two estimators correspond to the assumption that data follow a Gaussian distribution (either with the same scatter matrix for all batches or the same location).

Two \emph{Nearest centroïd classifiers} rely on the symmetrized KL divergence between Gaussian distributions.
\sloppy Let $\Mg = \R^p \times \Spos$. Given two pairs of parameters $\upsilon_1 = (\bMu_1, \bSigma_1) \in \Mg$ and $\upsilon_2 = (\bMu_2, \bSigma_2)$, this divergence is given by $\delta_\Mg(\upsilon_1, \upsilon_2) = \frac12( \delta_\text{KL}(\upsilon_1, \upsilon_2) + \delta_\text{KL}(\upsilon_2, \upsilon_1))$ where $\delta_\text{KL}(\upsilon_1, \upsilon_2) = \frac12 \left( \Tr\left(\bSigma_2^{-1} \bSigma_1 \right) + \Delta\bMu^T \bSigma_2^{-1} \Delta\bMu + \log\left( \frac{\Det{\bSigma_2}}{\Det{\bSigma_1}}\right) - p \right)$.
The center of mass of $\left\{\upsilon_i\right\}_{i=1}^M$ is the solution of $ \minimize_{\upsilon \in \Mg} \sum_{i=1}^M \delta_\Mg(\upsilon, \upsilon_i)$.
Then, two \emph{Nearest centroïd classifiers} are derived using two estimators: $\bSigmaGcentered$ and the MLE of the Gaussian distribution $(\bMuG, \bSigmaG)$.

Finally, the proposed \emph{Nearest centroïd classifier} on $\Mmsg$ relies on the symmetrized KL divergence~\eqref{eq:divergence} between NC-MSGs.
The center of mass is computed as explained in the subsection~\ref{subsec:center_of_mass} and the estimation is described in Section~\ref{sec:MLE} with the L2 penalty for the regularization.
For initialization, we used the arithmetic mean, \emph{i.e.} given a set of parameters $\{\theta_i \triangleq (\bMu_i, \bSigma_i, \bt_i) \}_{i=1}^M$, with $\theta_\textup{init} = \left(\bMu_\text{mean}, \bSigma_\text{mean}, N(\bt_\text{mean}) \right)$, where $\bMu_\text{mean} = \frac{1}{M} \sum_{i=1}^M \bMu_i$, $\bSigma_\text{mean} = \frac{1}{M} \sum_{i=1}^M \bSigma_i$, $\bt_\text{mean} = \frac{1}{M} \sum_{i=1}^M \bt_i$ and $N$ is the normalization function: $\forall \bx = (x_i)_{1\leq i \leq n} \in \RPosVec$, $N(\bx) = \left(\prod_{i=1}^n x_i \right)^{-\frac{1}{n}} \bx$.

The data are divided into two sets: a training set and a test set with $485\ 649$ and $122\ 614$ batches respectively~\cite{breizhcrops2020}. %\footnote{This division was proposed in~\cite{breizhcrops2020} to get two sets that come from two different geographical areas.}.
Among the six \emph{Nearest centroïd classifiers}, only the one on $\Mmsg$ has a hyperparameter which the parameter $\beta$ of the regularized NLL~\eqref{eq:NLLReg}.
Several values of $\beta$ are tested on a training set and a validation set, and both are subsets of the original training set.
The performance is measured with the ``F1 weighted" metric used in~\cite{breizhcrops2020} and is plotted in Figure~\ref{fig:num_exp:tuning_beta}.
The value of $\beta$ with the highest ``F1 weighted" metric is $10^{-11}$.
Hence, we use this value in the rest of the paper.
Then, we propose an experiment to illustrate Proposition~\ref{prop:minima_transformed_data} on the invariance of the estimation of textures under rigid transformations.
Indeed, we train the six \emph{Nearest centroïd classifiers} on a subset of the original training set and apply them to the full test set with a rigid transformation.
Thus, the more a \emph{Nearest centroïd classifier} is robust to these rigid transformations, the better the ``F1 weighted" metric.
Given $t\in [0, 1]$, three different rigid transformations are performed:
transformation of the mean $\bx_i \mapsto \bx_i + \bMu(t)$ with $\bMu(t) = t \,\ba$ for a given $\ba \in \R^p$,
rotation transformation $\bx_i \mapsto \bQ(t)^T\bx_i$ with $\bQ(t) = \exp(t \bXi)$ for a given skew-symmetric $\bXi \in \R^{p \times p}$ (hence $\bQ(t) \in \Op$),
and the joint mean and rotation transformation $\bx_i \mapsto \bQ(t)^T\bx_i + \bMu(t)$.
It should be noted that at $t=0$, the data are left unchanged.
The results are presented in Figure~\ref{fig:num_exp:rigid_transfo}.

The conclusions of these experiments are fourfold.
First, the proposed \emph{Nearest centroïd classifier} applies to large-scale datasets such as the \emph{Breizhcrops} dataset.
Second, the regularization proposed in Section~\ref{sec:MLE} is important to get good classification performance.
Indeed, we observe from Figure~\ref{fig:num_exp:tuning_beta} that if $\beta$ is too small, then the ``F1 weighted" metric becomes very low.
Also, if $\beta$ is too large, then the ``F1 weighted" metric also becomes very low.
Third, using KL divergences and their associated centers of mass to classify estimators gives much better performance than the classical Euclidean distance.
Indeed, even when data do not undergo rigid transformations, \emph{Nearest centroïd classifiers} based on KL divergences outperform Euclidean \emph{Nearest centroïd classifiers} in  Figure~\ref{fig:num_exp:rigid_transfo}.
Fourth, considering NC-MSGs, as well as its KL divergence, instead of the Gaussian distribution, is interesting to classify time series especially when rigid transformations are applied to the data.
Indeed, in  Figure~\ref{fig:num_exp:rigid_transfo}, we observe a large performance improvement when data are considered distributed from a NC-MSG and undergo rigid transformations.

\section{Conclusion}
	In this paper, we proposed a Riemannian gradient descent algorithm based on the Fisher-Rao information geometry of the NC-MSG.
	This algorithm is leveraged for two problems: parameter estimation and computation of centers of mass.
	The estimation problem of the NC-MSG is not straightforward. %has been little studied in the literature.
	Indeed, a major issue is that the existence of a solution to the NLL minimization problem is not guaranteed.
	To overcome this issue, we proposed a class of regularized NLLs that make the trade-off between a white Gaussian distribution and the NC-MSG.
	These functions are guaranteed to have a minimum, and this result holds without conditions on the samples.
	Furthermore, we derived the KL divergence between NC-MSGs which enabled us to define the centers of mass of NC-MSGs as minimization problems.
	The latter is solved using the proposed Riemannian gradient descent.
	Simulations have shown that the proposed  Riemannian gradient descent is fast on both minimization problems.
	Also, a \emph{Nearest centroïd classifier} based on the KL divergence has been implemented.
	It has been applied on the large-scaled dataset \emph{Breizhcrops} and showed robustness to transformations of the test set.

\clearpage
%\vfill\pagebreak

\bibliographystyle{IEEEbib}
\bibliography{refs}

\onecolumn
\newpage

%\allsectionsfont{\centering}
\section*{
	\Large Supplementary Material:\\
	Riemannian optimization for a non-centered mixture of scaled Gaussian distributions \\ \, \\
	\normalsize Antoine Collas, Arnaud Breloy, Chengfang Ren, Guillaume Ginolhac, Jean-Philippe Ovarlez \\ \, \\
}
\appendix
\label{sec:supp_material}

% Full supplementary
% \input{./sections/7_supplementary.tex}

% Compact supplementary
\begin{multicols}{2}
	\subsection{Proof of Proposition~\ref{prop:FIM}: Fisher Information Metric}
\label{supp:subsec:proof_FIM}

First, we recall the definition of the FIM.
See~\cite{S05} for an in-depth presentation.
Let $\{\bx_i\}_{i=1}^n$ be data points.
Assuming that the underlying distribution admits a p.d.f., the corresponding NLL, denoted $\NLL$, maps parameters $\theta$ belonging to the parameter space of the p.d.f., denoted $\M$, onto $\R$.
By denoting $T_\theta \M$ the tangent space of $\M$ at $\theta \in \M$, and under conditions of regularity of $\NLL$, the FIM is defined $\forall \xi, \eta \in T_\theta \M$ as
\begin{equation*}
	\langle \xi, \eta \rangle_{\theta}^{\M} =
	\E[\D \NLL(\theta)[\xi]\D \NLL(\theta)[\eta]] =
	\E[\D^2 \NLL(\theta)[\xi,\eta]].
\end{equation*}

To derive the FIM of the mixture of scaled Gaussians given in Proposition~\ref{prop:FIM}, we recall classical formulas for the Gaussian distribution.
The NLL at $\theta = (\bMu, \bSigma) \in \Mg = \R^p \times \Spos$ and associated to one data point $\bx$ is (neglecting terms not depending on $\theta$)
\begin{equation}
	\label{eq:NLL_G}
	\NLL_\bx^g(\theta) = \frac12 \left[\log |\bSigma| + (\bx - \bMu)^T \bSigma^{-1} (\bx - \bMu) \right]
\end{equation}
Since $\Mg$ is an open set in the vector space $\R^p \times \Sym$, the tangent space of $\Mg$ at $\theta$ is $T_\theta\Mg = \R^p \times \Sym$.
Finally, $\forall \xi=(\bXi_\bMu, \bXi_\bSigma), \eta=(\bEta_\bMu, \bEta_\bSigma) \in T_\theta \Mg$, the FIM of the Gaussian distribution associated to the NLL~\eqref{eq:NLL_G} is~\cite{S84}
\begin{equation}
	\label{eq:FIM_G}
	\langle \xi, \eta \rangle_{\theta}^\Mg = \bXi_\bMu^T\bSigma^{-1}\bEta_\bMu + \frac12 \Tr ( \bSigma^{-1} \bXi_\bSigma \bSigma^{-1}\bEta_\bSigma ).
\end{equation}

Then, we derive the FIM associated with the NLL of the mixture of scaled Gaussian distributions~\eqref{eq:NLL}.
We begin by writing~\eqref{eq:NLL} as a sum of Gaussian NLL~\eqref{eq:NLL_G}.
Indeed, $\forall \theta \in \Mmsg$, we have
\begin{equation*}
	\NLL(\theta | \{ \bx_i \}_{i=1}^n ) = \sum_{i=1}^n (\NLL_{\bx_i}^g \circ \varphi_i) (\theta),
\end{equation*}
where $\varphi_i(\theta) = (\bMu, \tau_i \bSigma)$.
Thus, $\forall \theta \in \Mmsg$, $\forall \xi, \eta \in T_\theta\Mmsg$, and following the reasoning of~\cite[Proposition 6]{BBGRP20} and~\cite[Proposition 3.1]{BMZSGB20}, the FIM of the mixture of scaled Gaussian is expressed as a sum of FIM of the Gaussian distribution~\eqref{eq:FIM_G}
\begin{align*}
	&\langle \xi, \eta \rangle_{\theta}^{\Mmsg} = \E \left[\D^2 \NLL(\theta | \{\bx_i\}_{i=1}^n)[\xi, \eta]\right] \\
							  &= \sum_{i=1}^n \E\left[\D^2 (\NLL_{\bx_i}^g \circ \varphi_i) (\theta)[\xi, \eta]\right] \\
							  &= \sum_{i=1}^n \E\left[\D (\NLL_{\bx_i}^g \circ \varphi_i) (\theta)[\xi] \D (\NLL_{\bx_i}^g \circ \varphi_i) (\theta)[\eta]\right] \\
							  &
							  \begin{multlined}
								  =\sum_{i=1}^n \E\big[\D (\NLL_{\bx_i}^g(\varphi_i(\theta)))[\D \varphi_i(\theta)[\xi]]\\ \D (\NLL_{\bx_i}^g(\varphi_i(\theta)))[\D \varphi_i(\theta)[\eta]]\big]
							  \end{multlined} \\
							  &= \sum_{i=1}^n \langle \D \varphi_i(\theta)[\xi], \D \varphi_i(\theta)[\eta]\rangle^{\Mg}_{\varphi_i(\theta)}.
\end{align*}
In the following, the $i$-th components of $\bXi_\bt$ and $\bEta_\bt$ are denoted $\xi_i$ and $\eta_i$ respectively.
Therefore, the directional derivative of the function $\varphi_i$ is
\begin{equation*}
	\D \varphi_i(\theta)[\xi] = (\bXi_\bMu, \xi_i \bSigma + \tau_i \bXi_\bSigma).
\end{equation*}
Thus, we get
\begin{align*}
	&\begin{multlined}
		\langle \xi, \eta \rangle_{\theta}^{\Mmsg} = \sum_{i=1}^n \big[\bXi_\bMu^T(\tau_i\bSigma)^{-1}\bEta_\bMu \\ + \frac12 \Tr\left((\tau_i \bSigma)^{-1} (\xi_i \bSigma + \tau_i \bXi_\bSigma) (\tau_i \bSigma)^{-1}  (\eta_i \bSigma + \tau_i \bEta_\bSigma)\right)\big]
	\end{multlined}\\
	&\begin{multlined}
		= \sum_{i=1}^n \Big[ \frac{1}{\tau_i}\bXi_\bMu^T\bSigma^{-1}\bEta_\bMu + \frac12 p \frac{\xi_i \eta_i}{\tau_i^2} + \frac12 \frac{\xi_i}{\tau_i} \Tr(\bSigma^{-1}\bEta_\bSigma) \\ + \frac12 \frac{\eta_i}{\tau_i} \Tr(\bSigma^{-1}\bXi_\bSigma) + \frac12 \Tr(\bSigma^{-1} \bXi_\bSigma \bSigma^{-1} \bEta_\bSigma) \Big]
	\end{multlined}\\
	&  
	\label{eq:FIM_last_calculus}
	\begin{multlined} 
		= \sum_{i=1}^n \left(\frac{1}{\tau_i}\right) \bXi_\bMu^T\bSigma^{-1}\bEta_\bMu + \frac{n}{2} \Tr(\bSigma^{-1} \bXi_\bSigma \bSigma^{-1} \bEta_\bSigma) \\ + \frac{p}{2} \left(\bXi_\bt \odot \bt^{-1}\right)^T \left(\bEta_\bt \odot \bt^{-1}\right) \\ + \frac12 \bXi_\bt^T \bt^{\odot -1} \Tr(\bSigma^{-1}\bEta_\bSigma) + \frac12 \bEta_\bt^T \bt^{\odot -1} \Tr(\bSigma^{-1}\bXi_\bSigma)
	\end{multlined}
\end{align*}
Since $\bXi_\bt, \bEta_\bt \in T_\theta \Mmsg$, we have $\bXi_\bt^T\bt^{\odot -1} = \bEta_\bt^T \bt^{\odot -1} = 0$.
Thus, the last two terms of the last equation cancel, and the expression of the FIM from Proposition~\ref{prop:FIM} is obtained
\begin{equation*}
	\begin{multlined}
		\langle \xi, \eta \rangle_{\theta}^{\Mmsg} = \sum_{i=1}^n \left(\frac{1}{\tau_i}\right) \bXi_\bMu^T\bSigma^{-1}\bEta_\bMu + \frac{n}{2} \Tr\left(\bSigma^{-1} \bXi_\bSigma \bSigma^{-1} \bEta_\bSigma\right)\\ + \frac{p}{2} \left(\bXi_\bt \odot \bt^{-1}\right)^T \left(\bEta_\bt \odot \bt^{-1}\right).
	\end{multlined}
\end{equation*}
It should be noted that this formula defines an inner product on $\Emsg$ if a transpose is added to $\bXi_\bSigma$.
Thus, $\langle ., . \rangle_.^\Mmsg$ is extended $\forall \xi, \eta \in \Emsg$ as presented in  Proposition~\ref{prop:FIM}.

\subsection{Proof of Proposition~\ref{prop:orth_proj}: orthogonal projection on \texorpdfstring{$\Mmsg$}{Mmsg}}
\label{supp:subsec:proof_orth_proj}

First of all, $\forall \theta \in \Mmsg$ the ambient space $\Emsg$ defined in~\eqref{eq:ambient_space} is decomposed into two complementary subspaces
\begin{equation}
	\label{eq:Emsg_decomp}
	\Emsg = T_\theta \Mmsg + T_\theta^\perp \Mmsg
\end{equation}
where $T_\theta \Mmsg$ is the tangent space at $\theta$ defined in~\eqref{eq:tangent_space} and $T_\theta^\perp \Mmsg$ is the orthogonal complement
\begin{equation}
	\label{eq:def_orth_complement}
	T_\theta^\perp\Mmsg \! = \! \left\{\xi \in \Emsg:\langle \xi, \eta \rangle_\theta^\Mmsg \! = \! 0, \forall \eta \in T_\theta \Mmsg\right\}.
\end{equation}
It can be checked that this orthogonal complement is
\begin{equation}
	\label{eq:orth_complement}
	T_\theta^\perp\Mmsg = \left\{\VEC{0}\right\} \times \Skew \times \left\{\alpha \bt : \alpha \in \RPos\right\}
\end{equation}
where $\Skew$ is the set of $p \times p$ skew-symmetric matrices.
Indeed, the elements of \eqref{eq:orth_complement} verify the definition~\eqref{eq:def_orth_complement} and $\dim(\Emsg) = \dim(T_\theta \Mmsg) + \dim(T_\theta^\perp \Mmsg)$.
Using the equations~\eqref{eq:Emsg_decomp} and~\eqref{eq:orth_complement}, the orthogonal projection of $\xi = (\bXi_\bMu, \bXi_\bSigma, \bXi_\bt) \in \Emsg$ onto $T_\theta\Mmsg$ is
\begin{equation*}
	P_\theta^\Mmsg (\xi) = \left( \bXi_\bMu, \bXi_\bSigma - \bA, \bXi_\bt - \alpha\bt \right)
\end{equation*}
where $\bA \in \Skew$ and $\alpha \in \RPos$ have to be determined.
% where $\sym(\bXi_\bSigma) = \frac12 (\bXi_\bSigma + \bXi_\bSigma^T)$ and $\alpha \in \RPos$ has to be determined.
Furthermore, $\forall \eta = \left( \VEC{0}, \bEta_\bSigma, \beta \bt \right) \in T_\theta^\perp \Mmsg$ with $\beta \in \RPos$, we must have
\begin{equation*}
	\langle P_\theta^\Mmsg(\xi), \eta \rangle_\theta^\Mmsg = 0.
\end{equation*}
This induces that 
\begin{equation*}
	\begin{cases}
		\bXi_\bSigma - \bA = \sym(\bXi_\bSigma)\\
		\alpha = \frac{\bXi_\bt^T \bt^{\odot -1}}{n}
	\end{cases}
\end{equation*}
where $\sym(\bXi_\bSigma) = \frac12\left(\bXi_\bSigma + \bXi_\bSigma^T\right)$.
Thus, we get the orthogonal projection from Proposition~\ref{prop:orth_proj}.

\subsection{Proof of Proposition~\ref{prop:LVconnection}: Levi Civita connection on \texorpdfstring{$\Mmsg$}{Mmsg}}
\label{supp:subsec:proof_LVconnection}

First of all, the FIM defined in Proposition~\ref{prop:FIM} is rewritten with a function $g$.
Indeed, let $\theta \in \Mmsg$ and $\xi, \eta$ be smooth vector fields on $\Mmsg$, the function $g$ is defined as
\begin{equation}
	g_\theta(\xi, \eta) = \langle \xi, \eta \rangle_\theta^\Mmsg.
\end{equation}
This function $g$ is of primary importance for the development of the Levi Civita connection.

We briefly introduce the Levi-Civita connection.
The general theory of it can be found in \cite[Ch. 5]{AMS08}.
The Levi-Civita connection, simply denoted $\nabla: (\xi, \eta) \mapsto \nabla_\xi\eta$, is characterized by the Koszul formula.
Let $\nu$ be a smooth vector field on $\Mmsg$, in our case the Koszul formula writes
\begin{equation}
	\label{eq:Koszul}
	\begin{multlined}
		g_\theta(\nabla_\xi\eta,\nu) - g_\theta(\D\eta[\xi],\nu) = \\ \frac12 \big(\D g_\theta [\xi](\eta, \nu) +  \D g_\theta [\eta] (\xi, \nu) - \D g_\theta[\nu] (\xi,\eta) \big)
	\end{multlined}
\end{equation}
where $\D g_\theta[\nu] (\xi,\eta)$ is the directional derivative of the function $g_\cdot(\xi,\eta): \theta \mapsto g_\theta(\xi,\eta) $.
We begin by computing  $\D g_\theta[\nu] (\xi,\eta)$:
\begin{equation*}
	% \label{eq:LC_step_1}
	\begin{aligned}
		- \D g_\theta[\nu] &(\xi,\eta) = \sum_{i=1}^n \bigg(\frac{\nu_i}{\tau_i^2}\bigg) \bXi_\bMu^T\bSigma^{-1}\bEta_\bMu\\
		& + \sum_{i=1}^n \bigg(\frac{1}{\tau_i}\bigg) \bXi_\bMu^T \bSigma^{-1} \bNu_\bSigma \bSigma^{-1} \bEta_\bMu \\
		& + n \Tr \left( \bSigma^{-1} \sym (\bXi_\bSigma \bSigma^{-1}\bEta_\bSigma) \bSigma^{-1} \bNu_\bSigma\right) \\
		& + p\left(\bXi_\bt \odot \bEta_\bt \odot \bt^{\odot -2}\right)^T \left(\bNu_\bt \odot \bt^{\odot -1}\right).
	\end{aligned}
\end{equation*}
Since the objective is to identify $\nabla_\xi\eta$ using~\eqref{eq:Koszul} and the FIM from Proposition~\ref{prop:FIM}, the last equation needs to be rewritten.
To do so, the following two terms are rewritten
\begin{multline*}
	\sum_{i=1}^n \bigg(\frac{\nu_i}{\tau_i^2}\bigg) \bXi_\bMu^T\bSigma^{-1}\bEta_\bMu = \\ p \left(\frac1p \bXi_\bMu^T\bSigma^{-1}\bEta_\bMu \1 \odot \bt^{\odot -1} \right)^T \left(\bNu \odot \bt^{\odot -1}\right),
\end{multline*}
and, since $\bNu_\bSigma \in \Sym$
\begin{multline*}
	\sum_{i=1}^n \bigg(\frac{1}{\tau_i}\bigg) \bXi_\bMu^T \bSigma^{-1} \bNu_\bSigma \bSigma^{-1} \bEta_\bMu = \\ \sum_{i=1}^n \bigg(\frac{1}{\tau_i}\bigg) \Tr \left(\bSigma^{-1} \sym(\bEta_\bMu \bXi_\bMu^T) \bSigma^{-1} \bNu_\bSigma \right).
\end{multline*}
Hence, we get the equation~\eqref{eq:LC_step_2}.
We then compute $\D g_\theta [\xi](\eta, \nu)$ in equation~\eqref{eq:LC_step_3}.
Using \eqref{eq:LC_step_2} and \eqref{eq:LC_step_3}, we calculate the right-hand side of the Koszul formula~\eqref{eq:Koszul} in equation~\eqref{eq:LC_step_4}.
By identification of the Koszul formula~\eqref{eq:Koszul} and orthogonal projection onto the tangent space, we get the Levi Civita connection from the Proposition~\ref{prop:LVconnection}.
\begin{figure*}[t]
	\begin{equation}
		\label{eq:LC_step_2}
		\begin{aligned}
			- \D g_\theta[\nu] (\xi,\eta) &= \sum_{i=1}^n \bigg(\frac{1}{\tau_i}\bigg) \Tr \left(\bSigma^{-1} \sym(\bEta_\bMu \bXi_\bMu^T) \bSigma^{-1} \bNu_\bSigma \right) + n \Tr \left( \bSigma^{-1} \sym\left(\bXi_\bSigma \bSigma^{-1}\bEta_\bSigma\right) \bSigma^{-1} \bNu_\bSigma\right) \\
			& \quad + p \left(\frac1p \bXi_\bMu^T\bSigma^{-1}\bEta_\bMu \1 \odot \bt^{\odot -1} \right)^T \left(\bNu_\bt \odot \bt^{\odot -1}\right) + p\left(\bXi_\bt \odot \bEta_\bt \odot \bt^{\odot -2}\right)^T \left(\bNu_\bt \odot \bt^{\odot -1}\right) \\
			& = n \Tr \left( \bSigma^{-1} \sym \bigg[ \frac1n \sum_{i=1}^n \frac{1}{\tau_i} \bEta_\bMu \bXi_\bMu^T +  \bXi_\bSigma \bSigma^{-1}\bEta_\bSigma \bigg] \bSigma^{-1} \bNu_\bSigma \right) \\
			& \quad + p \left( \bigg[\frac1p \bXi_\bMu^T\bSigma^{-1}\bEta_\bMu \1 + \bXi_\bt \odot \bEta_\bt \odot \bt^{\odot -1} \bigg] \odot \bt^{\odot -1} \right)^T \left(\bNu_\bt \odot \bt^{\odot -1}\right)
		\end{aligned}
	\end{equation}
	\begin{equation}
		\begin{aligned}
			\label{eq:LC_step_3}
			\D g_\theta [\xi](\eta, \nu) &= - \sum_{i=1}^n \bigg(\frac{\xi_i}{\tau_i^2}\bigg)  \bEta_\bMu^T\bSigma^{-1}\bNu_\bMu - \sum_{i=1}^n \bigg(\frac{1}{\tau_i}\bigg) \bEta_\bMu^T \bSigma^{-1} \bXi_\bSigma \bSigma^{-1} \bNu_\bMu - n \Tr ( \bSigma^{-1} \bEta_\bSigma \bSigma^{-1}\bNu_\bSigma \bSigma^{-1} \bXi_\bSigma) \\
						     & \quad - p \left(\bEta_\bt \odot \bNu_\bt \odot \bt^{\odot -2}\right)^T \left(\bXi_\bt \odot \bt^{\odot -1}\right)\\
						     & = - \sum_{i=1}^n \bigg(\frac{1}{\tau_i}\bigg) \bEta_\bMu^T \left( \frac{\bXi_\bt^T\bt^{\odot -2}}{\sum_{i=1}^n\frac{1}{\tau_i}} \bI_p +\bSigma^{-1} \bXi_\bSigma \right)  \bSigma^{-1}\bNu_\bMu - n \Tr \left(\bSigma^{-1} \sym(\bXi_\bSigma \bSigma^{-1} \bEta_\bSigma) \bSigma^{-1}\bNu_\bSigma \right) \\
						     & \quad - p\left(\bXi_\bt \odot \bEta_\bt \odot \bt^{\odot -2}\right)^T \left(\bNu_\bt \odot \bt^{\odot -1}\right)
		\end{aligned}
	\end{equation}
	\begin{equation}
		\label{eq:LC_step_4}
		\begin{aligned}
			& \frac12 \left(\D g_\theta [\xi](\eta, \nu) + \D g_\theta [\eta] (\xi, \nu) - \D g_\theta[\nu] (\xi,\eta) \right) =\\
			& \quad \sum_{i=1}^n \left(\frac{1}{\tau_i}\right) \bigg[ - \frac12 \bigg[ \bEta_\bMu^T \left( \frac{\bXi_\bt^T\bt^{\odot-2}}{\sum_{i=1}^n\frac{1}{\tau_i}} \bI_p +\bSigma^{-1} \bXi_\bSigma \right) + \bXi_\bMu^T \left( \frac{\bEta_\bt^T\bt^{\odot-2}}{\sum_{i=1}^n\frac{1}{\tau_i}} \bI_p +\bSigma^{-1} \bEta_\bSigma \right) \Bigg] \Bigg]  \bSigma^{-1}\bNu_\bMu \\
			& \quad + \frac{n}{2} \Tr \left( \bSigma^{-1} \left[ \frac1n \sum_{i=1}^n \bigg(\frac{1}{\tau_i}\bigg) \sym(\bEta_\bMu \bXi_\bMu^T) -  \sym(\bXi_\bSigma \bSigma^{-1}\bEta_\bSigma) \right] \bSigma^{-1} \bNu_\bSigma\right) \\
			& \quad + \frac{p}{2} \left( \left[\frac1p \bXi_\bMu^T\bSigma^{-1}\bEta_\bMu \1 -\bXi_\bt \odot \bEta_\bt \odot \bt^{\odot -1} \right] \odot \bt^{\odot -1}\right)^T \left(\bNu_\bt \odot \bt^{\odot -1}\right)
		\end{aligned}
	\end{equation}
	\hrule
\end{figure*}

\subsection{Proof of Proposition~\ref{prop:grad}: Riemannian gradient on \texorpdfstring{$\Mmsg$}{Mmsg}}
\label{supp:subsec:proof_Riem_grad}

Let $f: \Mmsg \rightarrow \R$ be a smooth function and $\theta$ be a point in $\Mmsg$.
We present the correspondence between the Euclidean gradient of $f$ (which can be computed using automatic differentiation libraries such as Autograd~\cite{autograd} and JAX~\cite{jax}) and the Riemannian gradient associated with the FIM defined in  Proposition~\ref{prop:FIM}.
The Euclidean gradient $\grad f(\theta) = (\G_\bMu, \G_\bSigma, \G_\bt)$ of $f$ at $\theta\in \Mmsg$ is defined as the unique element in $\R^p \times \R^{p \times p} \times \R^n$ such that $\forall \xi \in \R^p \times \R^{p \times p} \times \R^n$
\begin{equation*}
	\D f(\theta) [\xi] = \langle \grad f(\theta), \xi \rangle_\theta = \G_\bMu^T \bXi_\bMu + \Tr \Big( \G_\bSigma^T\bXi_\bSigma \Big) + \G_\bt^T \bXi_\bt.
\end{equation*}
Then, the Riemannian gradient $\grad_\Mmsg f(\theta) = (\G_\bMu^\Mmsg, \G_\bSigma^\Mmsg, \G_\bt^\Mmsg)$ is defined as the unique element in $T_\theta\Mmsg$ such that $\forall \xi \in T_\theta\Mmsg$ % (see \cite[Ch. 3]{AMS08} for a complete definition of the gradient on a Riemannian manifold)
\begin{equation*}
	\D f(\theta) [\xi] = \langle \grad_\Mmsg f(\theta), \xi \rangle_\theta^\Mmsg.
\end{equation*}
Hence, $\forall \xi \in T_\theta\Mmsg$, we get that
\begin{align*}
	\D f(\theta) [\xi] &= \G_\bMu^T \bXi_\bMu + \Tr \Big( \G_\bSigma^T\bXi_\bSigma \Big) + \G_\bt^T \bXi_\bt\\
	& = \left(\sum_{i=1}^n \frac{1}{\tau_i}\right) \left( \left(\sum_{i=1}^n \frac{1}{\tau_i}\right)^{-1} \bSigma \G_\bMu\right)^T \bSigma^{-1} \bXi_\bMu\\
	& \quad + \frac{n}{2} \Tr \left( \bSigma^{-1} \left( \frac{2}{n} \bSigma \G_\bSigma \bSigma \right)^T \bSigma^{-1} \bXi_\bSigma \right) \\
	& \quad + \frac{p}{2} \left(\bt^{\odot{-1}} \odot \left( \frac{2}{p} \bt^{\odot 2} \odot \G_\bt \right) \right)^T \left(\bt^{\odot -1} \odot \bXi_\bt \right) \\
	& = \left(\sum_{i=1}^n \frac{1}{\tau_i}\right) \bEta_\bMu^T \bSigma^{-1} \bXi_\bMu + \frac{n}{2} \Tr \left( \bSigma^{-1} \bEta_\bSigma^T \bSigma^{-1} \bXi_\bSigma \right)\\
	& \quad + \frac{p}{2} \left(\bt^{\odot{-1}} \odot \bEta_\bt \right)^T \left(\bt^{\odot -1} \odot \bXi_\bt \right)
\end{align*}
where
\begin{align*}
	\eta & = \left( \bEta_\bMu, \bEta_\bSigma, \bEta_\bt \right)\\
	& = \left(\left(\sum_{i=1}^n \frac{1}{\tau_i}\right)^{-1} \bSigma \G_\bMu, \frac{2}{n} \bSigma \G_\bSigma \bSigma, \frac{2}{p} \bt^{\odot 2} \odot \G_\bt \right).
\end{align*}
To get the Riemannian gradient, it remains to project $\eta$ into the tangent space $T_\theta \Mmsg$ using the orthogonal projection $P_\theta^\Mmsg$.
Thus, we get the Riemannian gradient from the Proposition~\ref{prop:grad}.

\subsection{Proof of Proposition~\ref{prop:retr}: a second order retraction on \texorpdfstring{$\Mmsg$}{Mmsg}}
\label{supp:subsec:proof_retr}

\begin{figure*}[t]
	\begin{equation} \tag{46}
		\label{eq:second_diff_N}
 		\begin{aligned}
 			& \frac{d^2}{dt^2} (N \circ \bx)(t) = -\frac{1}{n} \left( \prod_{i=1}^n x_i(t) \right) \left( \bxtdot^T \bxt^{\odot -1} \right) \left( \prod_{i=1}^n x_i(t) \right)^{-\frac{1}{n}-1} \left[ \bxtdot - \frac{1}{n} \left( \bxtdot^T \bxt^{\odot -1} \right) \bxt \right] \\
 			& \quad + \left( \prod_{i=1}^n x_i(t) \right)^{-\frac{1}{n}} \bigg[ \bxtddot - \frac{1}{n} \left(\bxtddot^T\bxt^{\odot{-1}}\right) \bxt + \frac{1}{n}\left(\left(\bxtdot^{\odot 2}\right)^T\bxt^{\odot{-2}}\right)\bxt - \frac{1}{n} \left(\bxtdot^T\bxt^{\odot{-1}}\right)\bxtdot \bigg]\\
 			& = \frac{1}{n} \left( \prod_{i=1}^n x_i(t) \right)^{-\frac{1}{n}} \bigg[ n\bxtddot + \left(\left(\bxtdot^{\odot 2}\right)^T\bxt^{\odot{-2}} - \bxtddot^T\bxt^{\odot{-1}} \right)\bxt - 2\left(\bxtdot^T\bxt^{\odot{-1}}\right)\bxtdot \\
 			& \quad + \frac{1}{n}\left(\bxtdot^T\bxt^{\odot{-1}}\right)^2\bxt \bigg],
 		\end{aligned}
 	\end{equation}
 	where $\bxtddot = \frac{d^2}{dt^2}\bxt$.
	\begin{equation} \tag{47}
		\label{eq:btddot(0)}
 		\begin{aligned}
 			\btddot(0) &= \frac{d^2}{dt^2}\left(N \circ \left(\bt + t\bXi_\bt + \frac{t^2}{2} \left( \bXi_\bt^{\odot 2} \odot \bt^{\odot -1} - \frac1p \bXi_\bMu^T\bSigma^{-1}\bXi_\bMu\1 \right) \right) \right)\evaltzero \\
 				   &= \bXi_\bt^{\odot 2} \odot \bt^{\odot -1} - \frac1p \bXi_\bMu^T\bSigma^{-1}\bXi_\bMu\1 + \frac{1}{n} \bigg[\left(\bXi_\bt^{\odot 2}\right)^T\bt^{\odot{-2}} - \left(  \bXi_\bt^{\odot 2} \odot \bt^{\odot -1} - \frac1p \bXi_\bMu^T\bSigma^{-1}\bXi_\bMu\1 \right)^T\bt^{\odot{-1}} \bigg] \bt\\
 				   &= \bXi_\bt^{\odot 2} \odot \bt^{\odot -1} - \frac1p \bXi_\bMu^T\bSigma^{-1}\bXi_\bMu\1 + \frac{1}{np} \bXi_\bMu^T\bSigma^{-1}\bXi_\bMu \left(\1^T\bt^{\odot{-1}}\right) \bt.
 		\end{aligned}
 	\end{equation}
 	\hrule
\end{figure*}

Let $\theta \in \Mmsg$, $\xi \in T_\theta\Mmsg$ and $t \in [0, t_{max}[$ where $t_{max}$ is to be defined.
We denote $r(t) = R(t\xi)$ where $R$ is defined in Proposition~\ref{prop:retr}, \emph{i.e.}
\begin{align*}
	& r(t) = \Bigg( \bMu + t\bXi_\bMu + \frac{t^2}{2} \left[\frac{\bXi_\bt^T \bt^{\odot -2}}{\sum_{i=1}^n\frac{1}{\tau_i}} \bI_p + \bXi_\bSigma \bSigma^{-1}\right]\bXi_\bMu ,\\
	& \bSigma + t\bXi_\bSigma + \frac{t^2}{2} \left( \bXi_\bSigma \bSigma^{-1} \bXi_\bSigma -\frac1n \sum_{i=1}^n \left(\frac{1}{\tau_i}\right) \bXi_\bMu \bXi_\bMu^T \right),  \\
	& N\left(\bt + t\bXi_\bt + \frac{t^2}{2} \left(  \bXi_\bt^{\odot 2} \odot \bt^{\odot -1} - \frac1p \bXi_\bMu^T\bSigma^{-1}\bXi_\bMu\1 \right) \right) \Bigg),
\end{align*}
where $\forall \bx \in \RPosVec$, $N$ is defined as $N(\bx) = \left(\prod_{i=1}^n x_i \right)^{-1/n} \bx$.

The objective is to prove that $r$ is a second-order retraction on $\Mmsg$.
The different properties of the definition of a second-order retraction are verified in the following; see~\cite[Ch. 4 and 5]{AMS08} for a complete definition.

First of all, we define $t_{max}$ such that $r$ is a valid retraction.
Indeed, $r$ must respect some constraints of positivity,
\begin{align}
	\label{eq:retr_sigma_cond}
	\bSigma \! + \! t\bXi_\bSigma \! + \! \frac{t^2}{2} \left[ \bXi_\bSigma \bSigma^{-1} \bXi_\bSigma -\frac1n \sum_{i=1}^n \left(\frac{1}{\tau_i}\right) \bXi_\bMu \bXi_\bMu^T \right] \succ \0, \\
	\label{eq:retr_tau_cond}
	\bt \! + \! t\bXi_\bt \! + \! \frac{t^2}{2} \left[ \bXi_\bt^{\odot 2} \odot \bt^{\odot -1} - \frac1p \bXi_\bMu^T\bSigma^{-1}\bXi_\bMu\1 \right] > \0,
\end{align}
where for $\bA \in \Sym$, $\bA \succ \0$ means $\bA$ is positive definite and for $\bx \in \R^n$, $\bx > \0$ means the components of $\bx$ are strictly positive.
Of course,~\eqref{eq:retr_sigma_cond}~and~\eqref{eq:retr_tau_cond} are not necessarily respected depending on the value of $t$.
To define the value of $t_{max}$ such that \eqref{eq:retr_sigma_cond} and \eqref{eq:retr_tau_cond} are respected, we begin by studying the eigenvalues of the left side of~\eqref{eq:retr_sigma_cond}.
To do so, let $\lambda^-(\bA)$ be the smallest eigenvalue of $\bA$ and $\bSigma(t)$ be the left side of~\eqref{eq:retr_sigma_cond}.
Thus, we get that
\begin{align}
	& \lambda^- \left( \bSigma(t) \right) \geq \lambda^- \left(\bSigma\right) + t\lambda^- \left(\bXi_\bSigma\right) \nonumber\\
	& \quad + \frac{t^2}{2} \left[ \lambda^-\left( \bXi_\bSigma \bSigma^{-1} \bXi_\bSigma \right) -\frac1n \sum_{i=1}^n \left(\frac{1}{\tau_i}\right) \norm{\bXi_\bMu}_2^2 \right] \nonumber\\
	&\geq \lambda^- \left(\bSigma\right) + t\lambda^- \left(\bXi_\bSigma\right) - \frac{t^2}{2n} \sum_{i=1}^n \left(\frac{1}{\tau_i}\right) \norm{\bXi_\bMu}_2^2 .
	\label{eq:retr_min_eig_sigma}
\end{align}
A sufficient condition to satisfy~\eqref{eq:retr_sigma_cond} is that the right side of~\eqref{eq:retr_min_eig_sigma} is strictly positive.
This is achieved whenever $t$ is in $\left[0, t_1 \right[$ where $t_1$ is defined as followed
\begin{itemize}
	\item if $\bXi_\bMu \neq \0$, $t_1 = n\frac{\lambda^-(\bXi_\bSigma) + \sqrt{\Delta_1} }{\sum_{i=1}^n \left(\frac{1}{\tau_i}\right) \norm{\bXi_\bMu}_2^2} > 0$ and $\Delta_1 = \lambda^-(\bXi_\bSigma)^2 + \frac2n \lambda^-(\bSigma) \sum_{i=1}^n \left(\frac{1}{\tau_i}\right) \norm{\bXi_\bMu}_2^2$,
	\item if $\bXi_\bMu = \0$, $t_1 = \frac{\lambda^-(\bSigma)}{\left|\lambda^-(\bXi_\bSigma)\right|}>0$ for $\lambda^-(\bXi_\bSigma) < 0$, $t_1 = + \infty$ otherwise.
\end{itemize}
Lets denote the minimum value coordinate of $\bx \in \R^n$ by $(\bx)_{min}$.
Using the same reasoning as before, one can show that whenever $t$ is in $[0, t_2[$, where $t_2$ is defined in the following, \eqref{eq:retr_tau_cond} is satisfied.
\begin{itemize}
	\item If $\bXi_\bMu \neq \0$, $t_2 = p\frac{(\bXi_\bt)_{min} + \sqrt{\Delta_2} }{\norm{\bSigma^{-\frac12}\bXi_\bMu}_2^2} > 0$
    and $\Delta_2 = (\bXi_\bt)_{min}^2 + \frac2p (\bt)_{min} \norm{\bSigma^{-\frac12}\bXi_\bMu}_2^2$.
	\item If $\bXi_\bMu = \0$, $t_2 = \frac{(\bt)_{min}}{\left|(\bXi_\bt)_{min}\right|} > 0$ for $(\bXi_\bt)_{min} < 0$, $t_2 = + \infty$ otherwise.
\end{itemize}
Hence, we get $t_{max} = \min\{t_1, t_2\} > 0$ such that $\forall t \in [0, t_{max}[$, $r(t) \in \Mmsg$.

Then, to be a second-order retraction, it remains to check that the three following properties are respected,
\begin{equation}
	\label{eq:retr_cond}
	r(0) = \theta, \quad \rdot(0) = \xi, \quad \nabla_\rdot \rdot\evaltzero =  0,
\end{equation}
where $\rdot(t) = \frac{d}{dt}r(t)$ and $\nabla$ is the Levi-Civita connection defined in Proposition~\ref{prop:LVconnection}.
The first property is easily verified.
In the rest of the proof, the following notations are used: $r(t) = (\bMu(t), \bSigma(t), \bt(t))$, $\rdot(t) = (\bMudot(t), \bSigmadot(t), \btdot(t))$ and $\rddot(t) = (\bMuddot(t), \bSigmaddot(t), \btddot(t))$.

We verify the second property of~\eqref{eq:retr_cond} which is $\rdot(0) = \xi$.
It is readily check that $\bMudot(0) = \bXi_\bMu$ and $\bSigmadot(0) = \bXi_\bSigma$.
It remains to verify that $\btdot(0) = \bXi_\bt$.
Computing the derivative of $N$ (defined in Proposition~\ref{prop:LVconnection}) at a point $\bxt \in \RPosVec$, we get that
\begin{multline}
	\label{eq:diff_N}
	\frac{d}{dt}(N \circ \bx)(t) =\\ \left[\prod_{i=1}^n x_i(t)\right]^{-\frac{1}{n}} \left[ \bxtdot - \frac{\bxtdot^T \bxt^{\odot -1}}{n} \bxt \right],
\end{multline}
where $\bxtdot = \frac{d}{dt}\bxt$.
Using this derivative and the constraints $\prod_{i=1}^n \tau_i = 1$ and $\bXi_\bt^T \bt^{\odot -1} = 0$, the desired property is derived
\begin{multline*}
	\btdot(0) = \frac{d}{dt}\bigg(N \circ \bigg(\bt + t\bXi_\bt \\ + \frac{t^2}{2} \bigg( \bXi_\bt^{\odot 2} \odot \bt^{\odot -1} - \frac1p \bXi_\bMu^T\bSigma^{-1}\bXi_\bMu\1 \bigg) \bigg) \bigg)\evaltzero = \bXi_\bt.
\end{multline*}
It remains to check the third condition of~\eqref{eq:retr_cond}.
Using the first two conditions of~\eqref{eq:retr_cond}, we get that $\nabla_\rdot \rdot\evaltzero =  0$ if and only if
\begin{equation*}
	\left\{ \!\!\! \begin{array}{lcc}
			\bMuddot(0) = \left[\frac{\bXi_\bt^T \bt^{\odot -2}}{\sum_{i=1}^n\frac{1}{\tau_i}} \bI_p + \bXi_\bSigma \bSigma^{-1}\right]\bXi_\bMu, & \\ 
			\bSigmaddot(0)  = \bXi_\bSigma \bSigma^{-1} \bXi_\bSigma - \frac1n \sum_{i=1}^n \left(\frac{1}{\tau_i}\right) \bXi_\bMu \bXi_\bMu^T, & \\
			P_\bt^\SRPosVec\left(\btddot(0)\right) \!\! = \!\! P_\bt^\SRPosVec \big(\bXi_\bt^{\odot 2} \odot \bt^{\odot -1} \!\! - \!\! \frac1p \bXi_\bMu^T\bSigma^{-1}\bXi_\bMu\1 \big),
	\end{array} \right.
\end{equation*}
where, $\forall \bXi \in \R^n$, $P_\bt^\SRPosVec(\bXi) = \bXi - \frac{\bXi^T\bt^{\odot-1}}n \bt$.
It is readily checked that the first two conditions are met.
Thus, only the third condition remains to be verified.
To do so, we differentiate~\eqref{eq:diff_N} to get the second derivative of $N$ in equation~\eqref{eq:second_diff_N}.
Using this derivative and the constraints $\prod_{i=1}^n \tau_i = 1$ and $\bXi_\bt^T \bt^{\odot -1} = 0$, the expression of $\btddot(0)$ is derived in the equation~\eqref{eq:btddot(0)}.
Using the linearity of the projection $P_\bt^\SRPosVec$, \eqref{eq:btddot(0)} implies that
\begin{align*}
	P_\bt^\SRPosVec(\btddot(0)) & = P_\bt^\SRPosVec\left(\bXi_\bt^{\odot 2} \odot \bt^{\odot -1} - \frac1p \bXi_\bMu^T\bSigma^{-1}\bXi_\bMu\1\right) \\
				    & \quad + P_\bt^\SRPosVec\left(\frac{1}{np} \bXi_\bMu^T\bSigma^{-1}\bXi_\bMu \left(\1^T\bt^{\odot{-1}}\right) \bt \right).
\end{align*}
Finally, one can check that $\forall \alpha \in \R$, $P_\bt^\SRPosVec(\alpha \bt) = 0$.
Hence, we get the desired expression
\begin{align*}
	P_\bt^\SRPosVec(\btddot(0)) = P_\bt^\SRPosVec\left(\bXi_\bt^{\odot 2} \odot \bt^{\odot -1} - \frac1p \bXi_\bMu^T\bSigma^{-1}\bXi_\bMu\1\right),
\end{align*}
which completes the proof.

\end{multicols}

\end{document}